\documentclass[lettersize,journal]{IEEEtran}
\usepackage{amsmath,amsfonts}
\usepackage{algorithmic}
\usepackage{tabularx}
\usepackage{arydshln}
\usepackage{array}
\usepackage{textcomp}
\usepackage{stfloats}
\usepackage{url}
\usepackage{verbatim}
\usepackage{graphicx}
\usepackage{amssymb}
\usepackage{bbding}
\usepackage{tablestyles}
\usepackage{times}
\usepackage{subfigure}
\usepackage{cite}

\def\BibTeX{{\rm B\kern-.05em{\sc i\kern-.025em b}\kern-.08em
    T\kern-.1667em\lower.7ex\hbox{E}\kern-.125emX}}
\usepackage{balance}
\begin{document}
\title{Simple and Robust Loss Design for Multi-Label Learning with Missing Labels}
\author{Youcai Zhang, Yuhao Cheng, Xinyu Huang, Fei Wen, Rui Feng, Yaqian Li and Yandong Guo
\thanks{
This work was done when Y. Cheng and X. Huang were
interning at OPPO Research Institute. They contributed equally with Y. Zhang. F. Wen is the corresponding author.

Y. Zhang, Y. Li and Y. Guo are with the OPPO Research Institute, Shanghai 200032, China (e-mails: zhangyoucai@oppo.com, liyaqian@oppo.com; guoyandong@oppo.com).

Y. Cheng and F. Wen are with the School of Electronic Information and Electrical Engineering,
Shanghai Jiao Tong University, Shanghai, China (e-mails: cyh958859352@sjtu.edu.cn, wenfei@sjtu.edu.cn).

X. Huang and R. Feng are with the Academy for Engineering and Technology, Fudan University, Shanghai, China (e-mails: xinyuhuang20@fudan.edu.cn, fengrui@fudan.edu.cn).

}}


\maketitle

\begin{abstract}
Multi-label learning in the presence of missing labels~(MLML) is a challenging problem. Existing methods mainly focus on the design of network structures or training schemes, which increase the complexity of implementation. 
This work seeks to fulfill the potential of loss function in MLML without increasing the procedure and complexity.
Toward this end, we propose two simple yet effective methods via robust loss design based on an observation that a model can identify missing labels during training with a high precision. The first is a novel robust loss for negatives, namely the Hill loss, which re-weights negatives in the shape of a hill to alleviate the effect of false negatives. The second is a self-paced loss correction~(SPLC) method, which uses a loss derived from the maximum likelihood criterion under an approximate distribution of missing labels. 
Comprehensive experiments on a vast range of multi-label image classification datasets demonstrate that our methods can remarkably boost the performance of MLML and achieve new state-of-the-art loss functions in MLML. 
Codes are available at \textit{\urlstyle{}{https://github.com/xinyu1205/robust-loss-mlml}}.
\end{abstract}

\section{Introduction}

\IEEEPARstart{R}{ecognizing} multiple labels of a given image, also known as multi-label image recognition, is an important and practical computer vision task, as images are intrinsically multi-labeled, {\it{e.g.}}, containing multiple objects, conditions, or scene types. In fact, the single-label dataset ImageNet has a large portion being composed of images with multiple labels. It has been shown in~\cite{yun2021re} that re-labeling ImageNet~\cite{deng2009imagenet} from single-label to multi-labels can yield remarkable performance improvement.


A main challenge for multi-label learning is the difficulty in collecting ``full'' labels for supervised training. In the case of a large number of categories, it is difficult and even impractical to simultaneously annotate the full label set for each training image. As a result, images are usually annotated with partial labels and, hence, available training data is often plagued by false negatives due to miss-labeling. In this context, multi-label learning in the presence of missing labels~(MLML) has attracted much research attention~\cite{yu2014large,durand2019learning,huynh2020interactive,cole2021multi}.


Generally, learning in the presence of missing labels is not straightforward, since for a given negative label there are two possibilities, {\it{1)}}~either it is truly negative that the class corresponding to it does not exist in the image, {\it{2)}}~or it is truly positive but simply not identified by the labeling mechanism. 
Existing works on MLML mainly utilize label correlation or transition matrix to estimate the missing labels by modifying the network or training procedures~\cite{huynh2020interactive,chu2018deep,bucak2011multi}, which typically require extra complex architectures or training schemes.

Besides, missing labels can also be regarded as a branch of label noise, where noise-robust loss design has shown remarkable performance in single-label recognition tasks~\cite{liu2015classification,zhang2018generalized,ren2018learning,wang2019symmetric}. However, robust loss design remains underexplored in multi-label learning.

\begin{figure}[t]
  \centering
  \includegraphics[width=.45\textwidth]{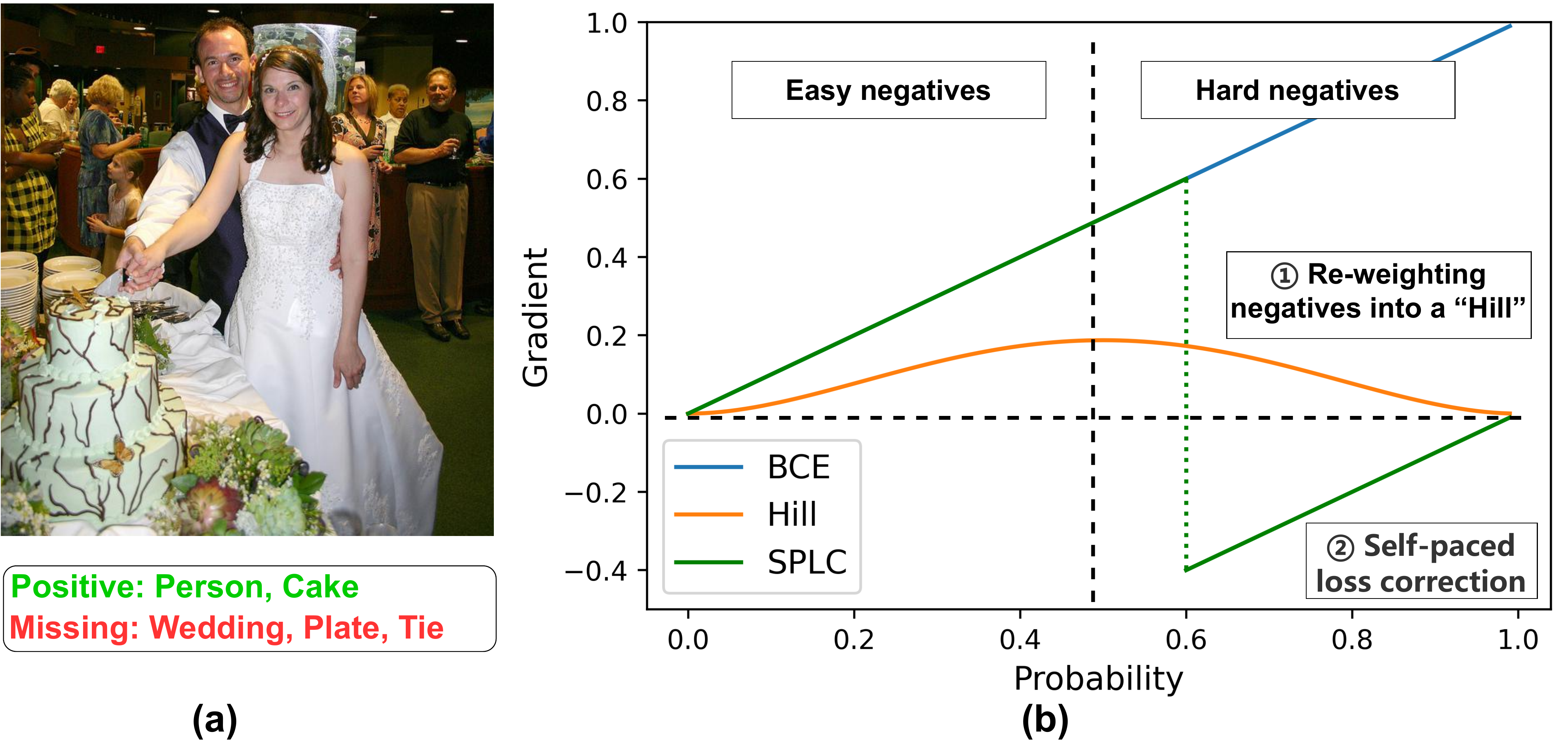}
  \caption{(a) Illustration of multi-label learning with missing labels. (b) Illustration of our proposed Hill loss and self-paced loss correction~(SPLC) for negatives.}
  \label{img:beginning}
\end{figure}

This work aims to provide some insights into MLML through investigating the efficacy of robust loss function. 
In particular, we demonstrate that a careful design of the loss function can greatly boost the robustness against the missing labels and the performance of classification accuracy, while still maintaining a simple and efficient solution, based on standard architectures and training schemes.
This benefits from the distinctive observation and characteristic in multi-label learning. 
We observe that the predicted probability from a model can be used to identify the false negatives~(missing labels) with a surprisingly high precision even at the early stage of training. 
Specifically, it is a consensus that there exists an imbalance between positives and negatives in multiple binary classifiers, which results in the domination of negatives on positives. Due to this fact, the predicted positives are inclined to be true positives~(including labeled positives and missing labels) with a high confidence. 



Inspired by this observation, we propose two simple yet effective methods to diminish the effect of missing labels. 
{\it{1)}}~The first is a new robust loss for negatives, namely the Hill loss, which is derived by re-weighting the MSE loss to alleviate the effect of false negatives. 
Specifically, the Hill loss puts less weight on potential false negatives and easy ones, but more weight on middle semi-hard ones to make the training of the model insensitive~(robust) to false negatives. 
{\it{2)}}~The second is a \textbf{S}elf-\textbf{P}aced \textbf{L}oss \textbf{C}orrection~(termed \textbf{SPLC}) method, which uses a loss derived from the maximum likelihood~(ML) criterion under an approximate probability distribution of the missing labels. 
Different from traditional correction techniques utilizing extra networks~\cite{chen2019multi,pineda2019elucidating} or extra training procedure~\cite{bucak2011multi}, the SPLC method gradually corrects missing labels based on model prediction in an automatically self-paced manner.


Moreover, we propose the Focal margin loss to exploit semi-hard positives by a simple modification to the Focal loss~\cite{lin2017focal}. 
Improvements brought by the Focal margin demonstrates that semi-hard mining is more appropriate than hard mining for positives. 
Extensive experiments show that our approaches can significantly outperform existing methods while do not introduce extra complex architectures or training schemes. 
The main contributions are summarized as follows:
\begin{itemize}
    \item An interesting observation in multi-label learning that a model can identify false negatives~(missing labels) even in the early stage of training;
    \item A systematic loss study on the problem of multi-label learning with missing labels that can simplify the pipeline in this field;
    \item A novel Hill loss to re-weight negatives and a simple self-paced loss correction method that are robust against missing labels;
    \item Comprehensive experiments that validate the effectiveness of our approaches and establish the new state-of-the-arts of loss function in MLML on multiple datasets.
\end{itemize}

\section{Related Work}
\subsection{Multi-label Learning with Missing Labels}
Multi-label learning with incomplete labels has been a hot topic in the community of weakly-supervised learning, due to the difficulty of annotating all the ground-truth labels~\cite{liu2021emerging}. 
There are two popular related research topics for incomplete labels, Partial Multi-Label Learning ~(PML)~\cite{xu2019partial, xie2021partial}, and Multi-label Learning with Missing Labels~(MLML)~\cite{yu2014large,durand2019learning}. 
To avoid the ambiguity of settings, we clarify the difference in Table~\ref{tab:missing kind}.
PML requires labeling all positive labels, which possibly introduces extra false positive labels and costs more. In other words, methods designed for PML typically rely on a candidate label set that includes all the positive labels occurring in the image.
By comparison, not all positive labels are required in MLML, which is more practical due to the lower labeling cost. 
In this work, we further relax the MLML setting, where no exact negative labels~\cite{durand2019learning} are utilized and the number of positive labels is not limited to single~\cite{cole2021multi}.
This means the scenario we are working on is more flexible and general in real-word scenarios.

\begin{table}[t]
\scriptsize
\caption{Comparison among different settings in multi-label classification. \Checkmark (resp. \XSolidBrush, \textcircled{}) means a label is present (resp. absent, unknown). Falsely marked labels are in red color. PML requires all positive labels are annotated. By comparison, MLML relaxes the requirement of annotations.}
\centering
\begin{tabular}{c|ccccc}
\hline
Settings &  label1 & label2 &  label3 & label4 & label5
 \\
\hline
Full labels & \Checkmark  &\Checkmark  &\XSolidBrush  &\XSolidBrush  &\XSolidBrush \\
PML &\Checkmark  &\Checkmark  &\textcolor{red}{\Checkmark}  &\textcircled{} &\textcircled{}\\
MLML &\Checkmark  &\textcolor{red}{\textcircled{}}  &\textcircled{}  &\textcircled{}  &\textcircled{}\\
\hline

\end{tabular}

\label{tab:missing kind}
\end{table}

Current works on MLML mainly focus on the design of networks and training schemes. The common practice is to utilize the customized networks to learn label correlations or classification confidence to realize correct recognition in missing labels.~\cite{zhang2018generalized} built a generative model to approximate the distributions of observed labels, by which the mislabeled samples can be recognized.~\cite{durand2019learning} proposed multi-stage training procedures to correct the missing labels. They utilized curriculum learning based strategies with the Bayesian uncertainty strategy for missing label correction. 

This domain has seen fast progress recently, at the cost of requiring more complex methods.
In contrast, this work seeks to fulfill the potential of the loss function in MLML, which is based on standard architectures, does not increase training and inference pipelines and time.
Besides, our methods are fully complementary to existing works and can be combined with them to further improve the performance at the cost of increasing the complexity.

\subsection{Noisy-robust Learning}
Noisy-robust learning is an important problem in training Deep Neural Networks~(DNNs) since label noise is ubiquitous in real-world scenarios.
The label noises can lead the DNNs to overfit to such noises due to the ``memorization effect'' proposed by \cite{zhang2021understanding,arpit2017closer}, eventually degrading the model generalization performance.
Sample re-weighting and loss correction are two typical approaches in this field. 

Sample re-weighting methods~\cite{liu2015classification,ren2018learning} typically reduce the contribution of noisy samples by down-weighting the loss of them. Many researchers~\cite{hu2019noise,zhong2019unequal} extended this idea to face recognition for noisy-robust feature learning.

Loss correction refers to the modification of the loss to compensate for the incorrect guidance provided by noisy samples. 
A basic approach is the pseudo label~\cite{lee2013pseudo}, which utilizes a trained model to correct annotations and then retrain the model using the updated labels. 
In~\cite{arazo2019unsupervised}, researchers found that clean and noisy samples can be distinguished from the loss distribution in noisy single-label datasets. Inspired by this observation, they proposed a mixture distribution method to correct noisy labels.
\cite{cole2021multi} proposed ROLE to jointly train classifiers and label estimator, by which the true labels can be estimated and corrected in the training phase. 

Though many works about the loss design achieve notable success and push forward the research in noise-robust learning, most~(if not all) of the existing works focus on the single-label task and ignore the multi-label task. 
This work validates that typical noise-robust loss functions in the single-label task~(both sample re-weighting and loss correction) can also be efficiently applied in multi-label learning.



\section{Simple and Robust Loss Design for MLML}

In this section, we first review classic losses in multi-label learning. Then, we make an observation in the scenario of missing labels, based on which the Hill loss and the SPLC method are introduced to deal with the probably mislabeled negatives. Furthermore, we make a simple modification on the positive part of the Focal loss to highlight semi-hard positives.

\subsection{Preliminary}
Multi-label classification is usually transformed into a multiple binary classification problem, in which multiple binary classifiers are trained to predict one-vs.-rest for each label independently.
Given $K$ labels, networks predict the logits $x_i$ of the $i$-th label independently, then the probabilities are given by normalizing the logits with the sigmoid function as $p_{i} = \sigma(x_i) =  \frac{1}{1+e^{-x_i}}$. Let $y_i$ denote the label for the $i$-th class, the binary classification loss is generally given by
\begin{equation}
    \mathcal{L} = - \sum_{i=1}^K \left( y_i L^+_i + (1-y_i)L^-_i \right),
\end{equation}
where $L_i^+$ and $L_i^-$ stand for the positive and negative losses of $i$-th label, respectively. For simplicity, in the sequel we ignore the subscript $i$ in $L_i^+$ and $L_i^-$ to use $L^+$ and $L^-$.


The Binary Cross Entropy~(BCE) loss is the most popular loss function in multi-label classification, which is defined as
\begin{equation}
\begin{cases}
    &L_{BCE}^+ = \log(p)    \\
    &L_{BCE}^- = \log(1-p)
\end{cases}.
\end{equation}

Further, the Focal loss~\cite{lin2017focal} is proposed to address the problem of positive-negative imbalance and hard-mining, which is given by
\begin{equation}
\begin{cases}
    &L_{Focal}^+ = \alpha_+(1-p)^\gamma\log(p)    \\
    &L_{Focal}^- = \alpha_-p^\gamma\log(1-p)
\end{cases},
\end{equation}
where $\gamma$ is a focus parameter, and $\alpha_+$ and $\alpha_-$ are utilized to balance positives and negatives. By using $\gamma>0$, hard samples are delivered more attention.

More recently, ASL~\cite{ridnik2021asymmetric} is proposed to relieve positive-negative imbalance by operating differently on positives and negatives, which is defined as:

\begin{equation}
\begin{cases}
    &L_{ASL}^+ = (1-p_m)^{\gamma_+}\log(p_m)    \\
    &L_{ASL}^- = p_m^{\gamma_-}\log(1-p_m)
\end{cases},
\end{equation}

where $\gamma_+<\gamma_-$ and are focus parameters for positive and negative labels, respectively, and $p_m=\max(p-m,0)$. The probability margin $m \geq 0$ is a tunable hyper-parameter. The ASL loss reduces the weight of easy negatives via using $\gamma_+ < \gamma_-$, and discards negatives with low predicted probability via the $m$ shifted probability.
\subsection{An Observation in MLML} 

\begin{figure}[t]
	\subfigure{
    	\begin{minipage}[t]{0.5\linewidth}
    	\centering
    	\includegraphics[width=1.7in]{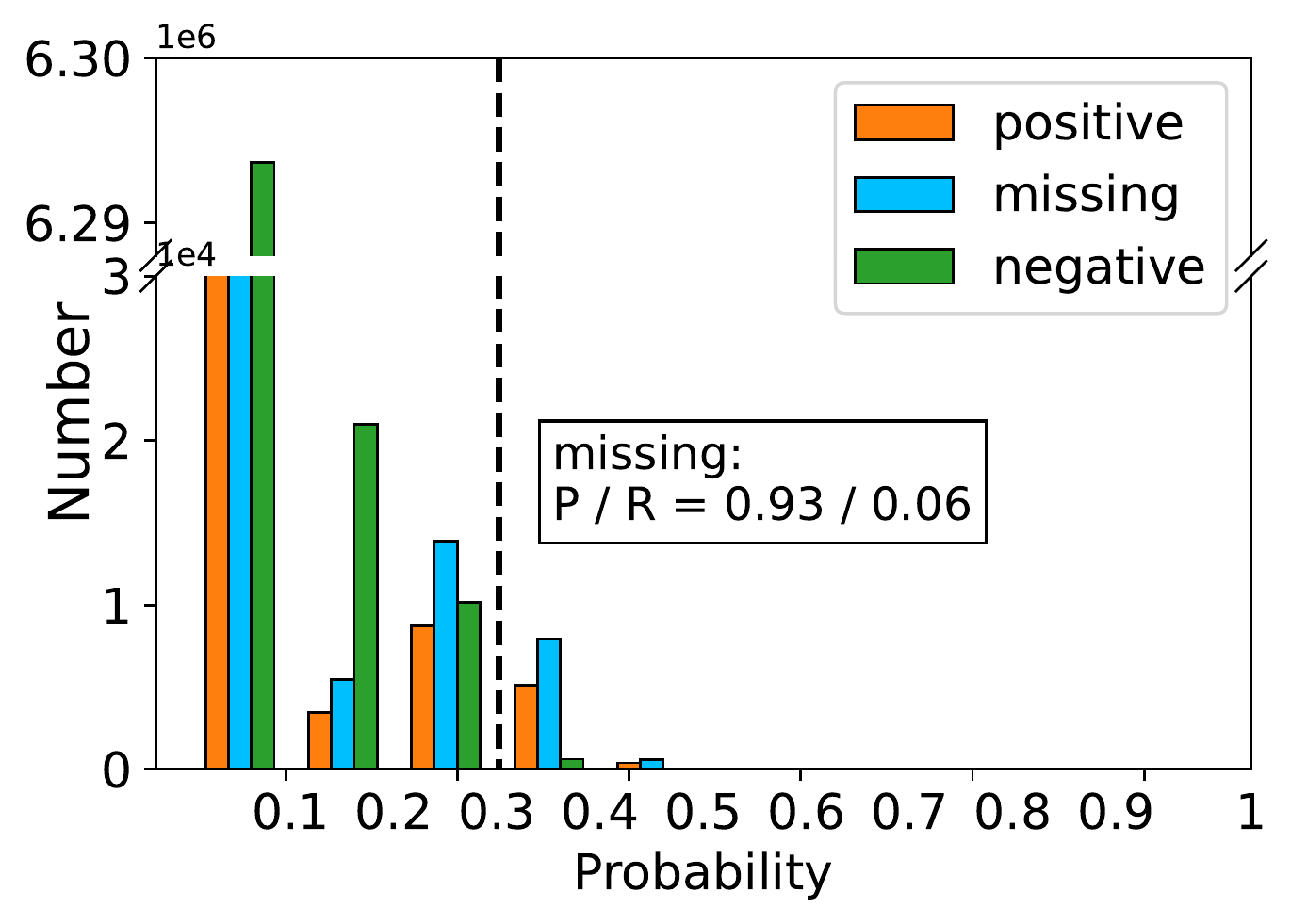}
    	\scriptsize (a)~The early stage of BCE
    	\centering
    	\includegraphics[width=1.7in]{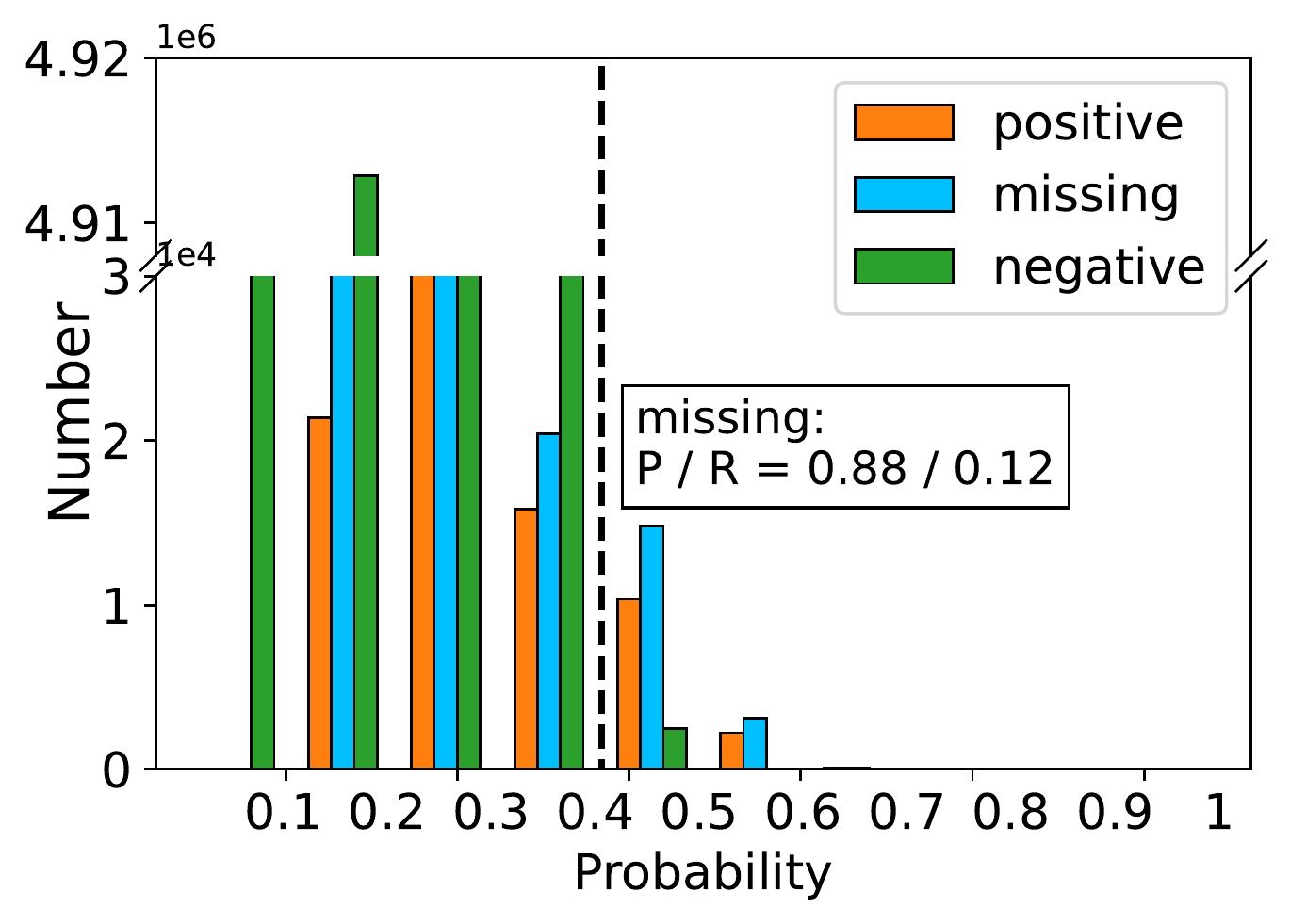}
	    \scriptsize (c)~The early stage of Focal
	    \end{minipage}%
	}%
	\subfigure{
    	\begin{minipage}[t]{0.5\linewidth}
    	\centering
    	\includegraphics[width=1.7in]{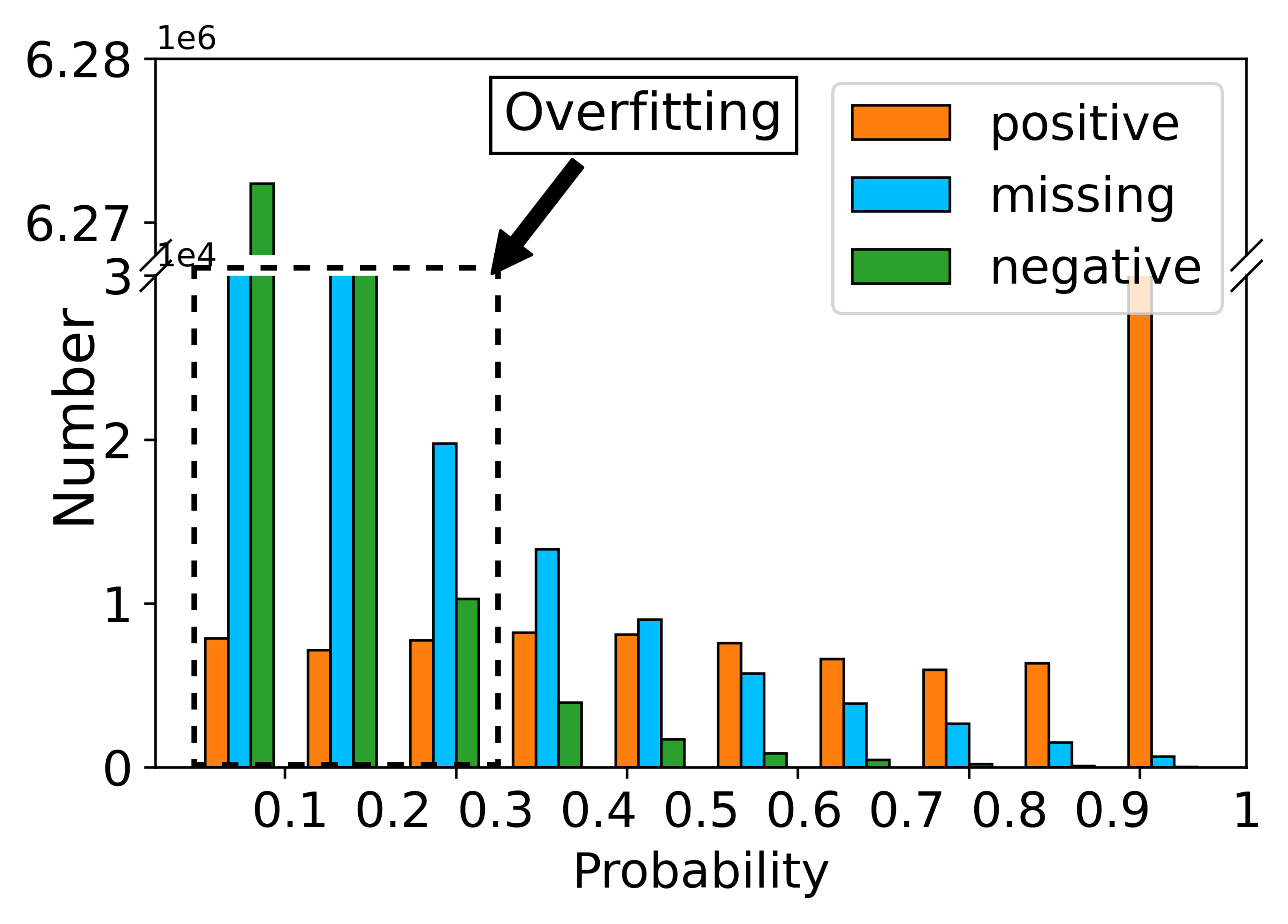}
    	\scriptsize (b)~The late stage of BCE
    	\centering
    	\includegraphics[width=1.7in]{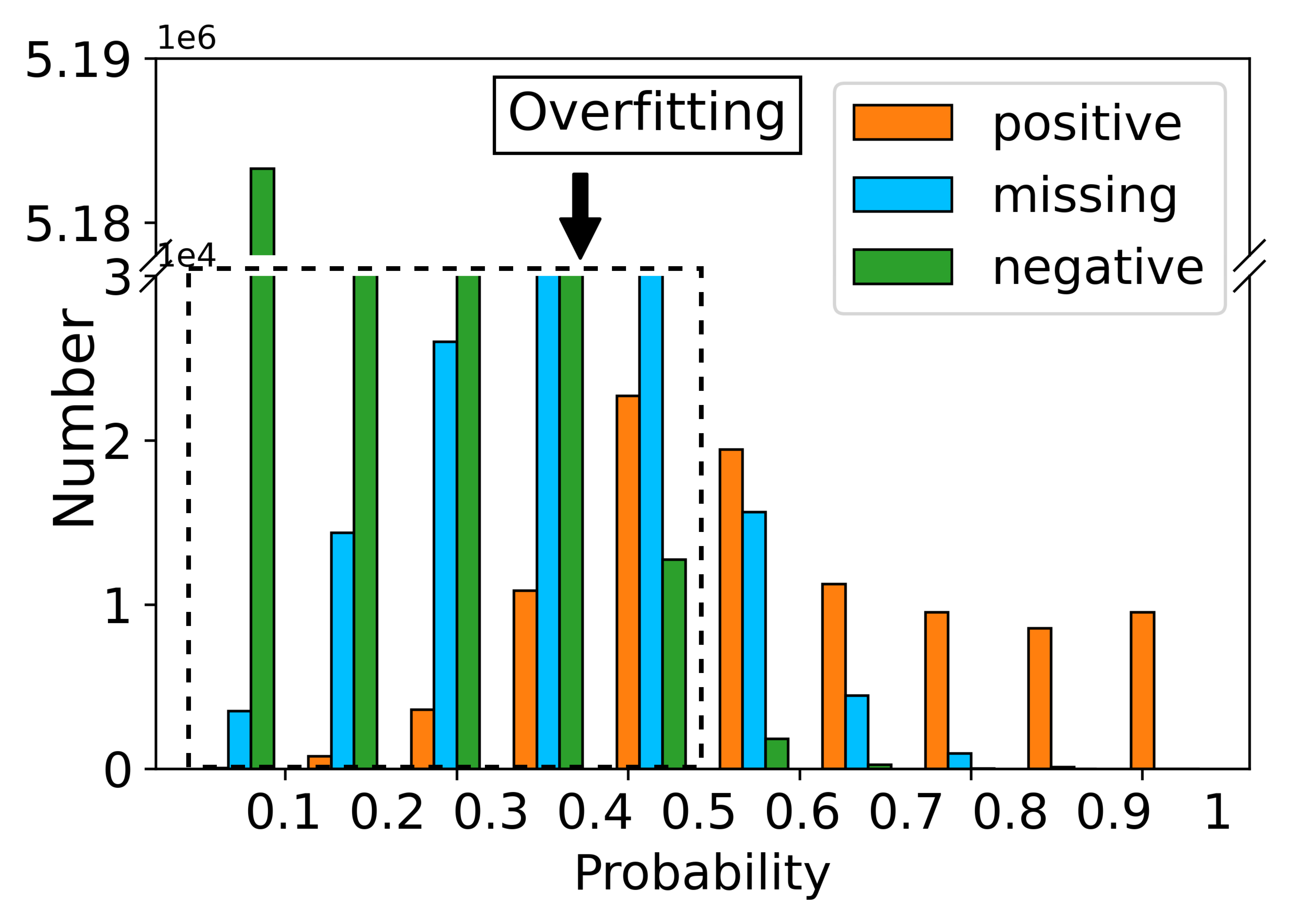}
    	\scriptsize (d)~The late stage of Focal
    	\end{minipage}%
	}%
	\centering
	\caption{Probability distribution of a network prediction at the early and late stages of training on the COCO training set with 40\% labels left. (a) and (c) indicate that false negatives can be identified with a high precision but a low recall at the early training stage. (b) and (d) show that the network predictions for missing labels share a similar distribution to that for true negatives, which implies that the network over-fits to labeled positives at the late stage of training.}
\label{img:observation}
\end{figure}

Under a typical missing label setting, Fig.~\ref{img:observation} shows the probability distribution of a network prediction at the early and late stages of training with BCE and Focal loss. We make an observation:
\emph{Missing labels~(false negatives,FNs) can be distinguished from true negatives~(TNs) according to the predicted probabilities of a network at the early training stage.}

Fig.~\ref{img:observation}(a) and (c) show that although most missing labels are poorly classified at the early training stage, partial missing labels can be identified with high precision but low recall by setting a proper threshold. For example, when using the BCE loss, partial FNs can be distinguished from TNs with a threshold of 0.3 on prediction probability. Similarly, when using the Focal loss, such a threshold can be set to 0.4. 


The reason behind this phenomenon is the distinctive features in multi-label learning: 
{\it{1)}} the domination of negatives on positives in multiple binary classifiers; In the multi-label setting, an image typically contains a few positives but much more negatives, which results in serious positive-negative imbalance~\cite{ben2020asymmetric}. 
{\it{2)}} Missing labels exacerbate the positive-negative imbalance and plague the learning of recognizing positives.
Therefore, the low ratio of positives make the model conservative in recognizing positives, once it makes such a prediction we can be confident it is correct.


Furthermore, it can be seen from Fig.~\ref{img:observation}(b) and (d) that, at the late stage of training, the network tends to over-fitting. As a result, the distribution of the network prediction for missing labels~(FNs) is similar to that for TNs, which makes the missing labels indistinguishable from true negatives.
That is, the network tends to prioritize learning simple patterns~(e.g., true labels) first before eventually memorizing~(over-fitting to) hard ones~(e.g., noisy/false labels).
This phenomenon accords well with the mainstream of viewpoint in noise-robust learning, \textit{memorization effect}, that ``deep neural networks easily fit random labels'' ~\cite{zhang2021understanding} and ``deep neural networks~(DNNs) learn simple patterns first, before memorizing''~\cite{arpit2017closer}. 
In light of this, it would be beneficial to exploit these distinguishable missing labels at the early stage of training rather than the late stage.

The above analysis sheds some light on how to relieve the effect of false negatives in the presence of missing labels. In the following, we propose two approaches via down-weighting and correcting potential missing labels, respectively. 


\label{sec_observation}

\subsection{Re-weighting Negatives into a ``Hill''}
\begin{figure}[t]
  \centering
  \includegraphics[width=.40\textwidth]{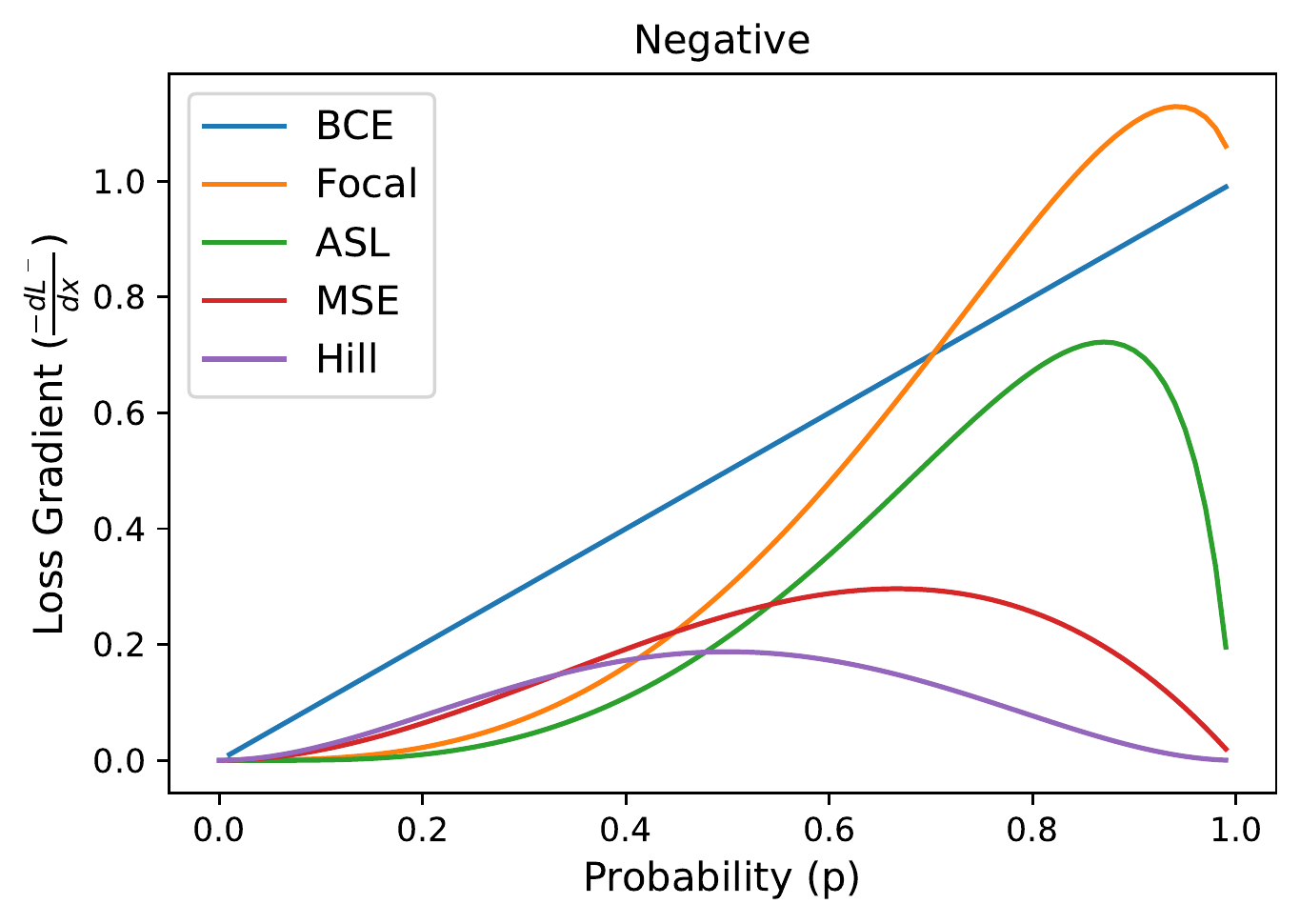}
  \caption{Gradient analysis of Hill loss~(shaped like a hill) in comparison with current loss functions. Hill loss puts more attention on semi-hard samples, while potential false negatives with high probability are delivered less emphasis.}
  \label{img:negloss}
\end{figure}

A straightforward method to alleviate the effect of missing labels is to design a robust loss that is insensitive to false negatives. Generally, the prediction probability of a well-trained model should be closer to 1 than 0 for false negatives. Hence, the effect of false negatives can be mitigated by putting less weight on the prediction probability closing to 1 in the loss for negative labels. As shown in Fig.~\ref{img:negloss}, the Mean Squared Error~(MSE) loss naturally satisfies this desirable robustness property, as it has a low weight on prediction probability closing to 1. Hence, it would be more robust than the BCE loss and can yield a better performance in the presence of missing labels. 

To further improve the robustness against false negatives, we propose the Hill loss by re-weighting the MSE loss as
\begin{equation}
\begin{aligned}
    \mathcal{L}_{Hill}^- &= -w(p)\times MSE\\
    &=-(\lambda-p){p}^2.
\end{aligned}
\end{equation}

The weighting term $w(p)=\lambda-p$ is designed to down-weight the loss for possibly false negatives. To fulfill this goal, considering a typical threshold of 0.5 for sigmoid-based binary classification, the hyper-parameter $\lambda$ can be selected to satisfy $\frac{\partial^2L_{Hill}^-}{\partial x^2}=0$ for $p=\sigma(x)=0.5$. Accordingly, the solution is given by $\lambda=1.5$, with which we obtain the Hill loss as illustrated in Fig.~\ref{img:negloss}. Clearly, it can be seen from Fig.~\ref{img:negloss} that the Hill loss has less weight than the MSE loss for $p\geq0.5$. Hence, it can be expected to achieve more robust performance against missing labels than MSE.

Fig.~\ref{img:negloss} compares the gradients of different loss functions for negative labels. It is noteworthy that each of the ASL, MSE and Hill losses has less weight on possible false negatives compared with the BCE loss. However, ASL can still be seriously affected by missing labels, since it only down-weights the region very close to 1, {\it{e.g.}}, $p > 0.9$. In fact, a large part of missing labels has a prediction probability less than 0.9, as shown in Fig.~\ref{img:observation}. Though the ASL loss can be adjusted by the parameters in Eq.~(4), the adjustable extent is restricted, which is analyzed in detail in Appendix B.

\subsection{Self-Paced Loss Correction}
This subsection presents a loss correction method to alleviate the effect of missing labels. It can gradually corrects the potential missing labels during training, so we term it \textbf{S}elf-\textbf{P}aced \textbf{L}oss \textbf{C}orrection~(\textbf{SPLC}).
SPLC can be viewed as an approximate maximum likehood~(ML) method, as it adopts a loss derived from the ML criterion under an approximate distribution of missing labels.

First, for the full label setting~(without missing labels), the BCE loss (Eq.~(2)) is optimal from the ML criterion under Bernoulli distribution
\begin{equation}
P_r(y) = p^y(1-p)^{1-y} = 
\begin{cases}
    p, &y=1\\
    1-p, & y=0
\end{cases}.
\end{equation}
Meanwhile, minimizing Eq.~(2) is equivalent to minimizing the KL divergence between the label and the model output. When in the presence of missing labels, for a negative label $y=0$, let $q\in[0,1]$ be the probability of its corresponding class being truly negative, while $1-q$ is the probability of its corresponding class being false negative~(caused by miss-labeling). 
In this setting, the distribution Eq.~(6) on longer applies and a modification is:
\begin{equation}
\begin{aligned}
P_r(y,s) &= (qp)^y {\left\{(1-p)^{s} [(1-q)p]^{1-s}  \right\}}^{1-y} \\ &= 
\begin{cases}
    qp, &y=1\\
    1-p, & y=0,s=1 \\
    (1-q)p, &y=0,s=0
\end{cases},
\end{aligned}
\end{equation}
where $s\in\{0,1\}$ is an indicating variable of true and false negatives, with $s=0$ standing for a false negative. 
With this distribution, the optimal loss from the ML criterion is given by
\begin{equation}
\begin{cases}
    &L^+ = \log(p)    \\
    &L^- = s\log(1-p) + (1-s)\log(p)
\end{cases}.
\end{equation}

Due to labeling mechanism, which negative labels are missing labels is not a prior known, {\it{i.e.}}, $s$ is not a prior known. Hence, the loss (Eq.~(8)) cannot be directly used for training. 
However, in practice the missing labels can be empirically identified based on the model prediction as analyzed in Sec.~\ref{sec_observation}. 
Hence, for a given negative label $y=0$, we can set a threshold $\tau\in(0,1)$ to identify whether it is a truly negative or false negative based on the model prediction probability $p$. Specifically, for a given negative label, it would be a missing label with high probability if $p>\tau$. In light of this understanding, a new loss that is robust to missing labels can be designed by recasting Eq.~(8) into:
\begin{equation}
\begin{cases}
    &L^+ = \log(p)    \\
    &L^- = \mathbb{I}(p\leq \tau)\log(1-p) + (1-\mathbb{I}(p\leq \tau))\log(p)
\end{cases},
\end{equation}
where $\mathbb{I}(\cdot)\in\{0,1\}$ is the indicator function. 

Furthermore, we can combine SPLC with the generalized loss function as
\begin{equation}
\begin{cases}
    &L_{SPLC}^+ = loss^+(p)    \\
    &L_{SPLC}^- = \mathbb{I}(p\leq \tau)loss^-(p) + (1-\mathbb{I}(p\leq \tau))loss^+(p)
\end{cases},
\end{equation}
where $loss^+(\cdot)$ and $loss^-(\cdot)$ refer to generalized loss functions for positives and negatives, respectively.


In implementation, SPLC adopts the label correction for the probability that exceeds a fixed threshold during training. It presents two main advantages compared with the pseudo label method: {\it{1)}}~SPLC is more efficient. In the pseudo-label method, a trained model is utilized to correct the labels, then the network is retrained. In contrast, the network only needs to be trained once in SPLC. {\it{2)}}~SPLC is more accurate. In the pseudo-label method, the network for label correction is trained to converge by noisy data, leading to poor capability in recognizing missing labels. In contrast, correcting missing labels with SPLC at the early stage of training reduces the influence of noisy data.

\begin{figure}[t]
  \centering
  \includegraphics[width=.40\textwidth]{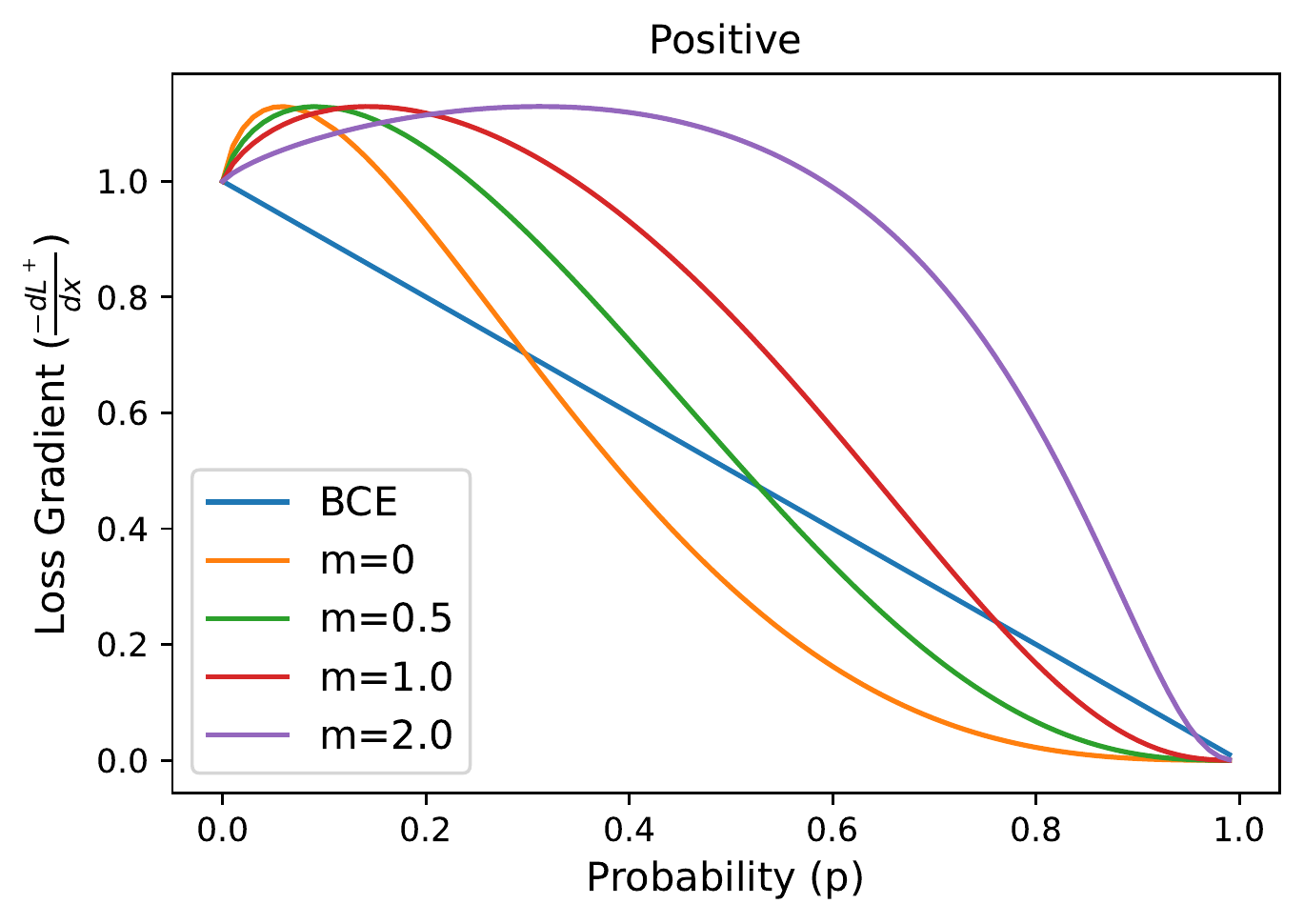}
  \vspace{-.5em}
  \caption{Gradient analysis of Focal margin loss with different margins in comparison with BCE. Without margin~($m=0$), Focal margin degrades into Focal loss, focusing on hard samples, whereas a proper margin~($m\in[0.5,1.0]$) enables Focal margin to highlight semi-hard samples.}
  \vspace{-1.5em}
  \label{img:posloss}
\end{figure}
\subsection{Semi-hard Mining for Positives}
In addition to dealing with negatives via Hill and SPLC, we also delicately tailor the loss for positives to exploit the feature of multi-label learning. 
In fact, hard-mining methods that can yield improvement in single-label learning are not widely used in multi-label learning. 
For example, the recent state-of-the-art method ASL only adopts the naive BCE loss for positives.

Here, we attempt to explain why hard-mining does not perform well in multi-label learning by the comparative analysis of Fig.~\ref{img:observation} (b) and (d).
It indicates that excessively hard mining brought by Focal loss is inappropriate for multi-label learning. 
Specifically, despite the decrease in the number of hard positives~(low probability, below $0.2$), Focal loss only pushes the proportion of 1.5\% hard positives to a high probability~(above 0.5).
Meanwhile, Focal loss ignores a large proportion of semi-hard ones with moderate probability~({\it{e.g.}}, in the region $[0.3, 0.5]$). 

Therefore, we seek to further emphasize semi-hard positives compared with Focal loss.
An intuitive approach is to subtract a margin from the logits. In this manner, they are regarded as smaller values and given larger gradients because of the feature of Focal loss. Meanwhile, thanks to the characteristic ``S''-shaped curve of sigmoid function, classes with probability in the middle region~(the logit value around 0, with the corresponding probability value around 0.5) are heavily affected,
meaning that they are given larger gradients compared with those at both ends~(easy and hard samples). 

Formally, we propose the Focal margin loss as:
\begin{equation}
    L^+_{Focal\;margin} = (1-p_{m})^\gamma\log(p_{m})
\end{equation}
where $p_{m} = \sigma(x-m)$ and $m$ is a margin parameter. Focal margin loss degrades to Focal loss when $m=0$. $\gamma$ is set to 2, a commonly used value in Focal loss.

The gradients of BCE and Focal margin losses with different $m$ are shown in Fig.~\ref{img:posloss}. It can be observed that Focal margin endows larger weight on semi-hard positives compared with Focal loss. In our work, $m\in[0.5,1.0]$ is recommended to avoid the highlight of easy positives. 

Actually, the introduction of margin is widely used in the design of loss function, most notably in SVMs~\cite{cortes1995support}. 
Recently, Large-Margin Softmax~\cite{liu2016large} and Arcface~\cite{deng2019arcface} have been proposed to minimize intra-class variation and at the meantime enlarge the inter-class margin by using the angular margin. This paper is the first work to perform semi-hard mining through the combination of margin and sigmoid activation.

To sum up, the Focal margin loss for positives are integrated with Hill loss and SPLC for negatives respectively into our final approaches~(denoted by Hill and Focal margin + SPLC). 

\section{Experiment}
In this section, we firstly introduce the experimental settings. Secondly, we compare proposed methods with existing classic loss functions on four multi-label datasets. 
Thirdly, ablation study and comprehensive analysis are given to delve into the working mechanism of Hill and SPLC.
Finally, more discussions and experiments are provided to answer the common questions in implementation.

\subsection{Experimental Settings}
\subsubsection{Datasets}
We conduct experiments on multiple standard multi-label benchmarks: MS-COCO~(COCO)~\cite{lin2014microsoft}, PASCAL VOC 2012~(VOC)~\cite{everingham2011pascal}, NUS-WIDE~(NUS)~\cite{chua2009nus}, and the large-scale Open Images~\cite{kuznetsova2020open}. COCO~\cite{lin2014microsoft} consists of 82,081 images with 80 classes for training and 40,137 images for test. VOC~\cite{everingham2011pascal} contains 5,717 training images with 20 classes, and additional 5,823 images are used for test.
Following \cite{ben2020asymmetric}, we collect NUS-wide~\cite{chua2009nus} with 119,103 and 50,720 images as training set and test set, consists of 81 classes.
Due to many download urls of Openimages~\cite{kuznetsova2020open} are expired, we were able to collect 1,742,125 training images and 37,306 test images, which contain 567 unique classes.

We construct the training sets of missing labels by randomly dropping positive labels for each training image with different ratios. 
The number of remaining labels per image $n_r(x)=\lfloor n(x) \times (1-r)\rfloor+1$, where $n(x)$ refers to the number of all positive labels per image and $r$ controls the ratio of missing labels. 
In particular, $r=1$ means single positive label per image, the same setting as~\cite{cole2021multi}. We report results for $r\in\{0.5, 0.8, 1\}$ respectively on COCO and single positive label ($r=1$) results on VOC, NUS and Open Images. 
Note that the test sets are fully labeled.
The detailed statistics of datasets are shown in Table~\ref{tab:dataset}.
\begin{table}[t]
\caption{The statistics of different datasets.}
\begin{center}
\setlength{\tabcolsep}{1mm}{
\begin{tabular}{c|cccc}
\hline
  &
\multicolumn{1}{c}{samples} &
\multicolumn{1}{c}{classes} &
\multicolumn{1}{c}{labels} &
\multicolumn{1}{c}{avg.label/img} 
 \\
  \hline
COCO-full labels  & 82,081 & 80 &241,035 &2.9 \\
COCO-75\% labels left  & 82,081 & 80 &181,422 &2.2 \\
COCO-40\% labels left  & 82,081 & 80 &96,251 &1.2 \\
COCO-single label  & 82,081 & 80 &82,081 &1.0 \\
\hline
NUS-full labels  & 119,103 & 81 &289,460 &2.4 \\
NUS-single label  & 119,103 & 81 &119,103 &1.0 \\
\hline
VOC-full labels  & 5,823 & 20 &8,331 &1.4 \\
VOC-single label  & 5,823 & 20 &5,823 &1.0 \\
\hline
Open Images-full labels & 1,742,125 & 567 & 4,422,289 & 2.54\\
Open Images-single label & 1,742,125 & 567 & 1,742,125 & 1  \\
\hline
\end{tabular}
}
\end{center}

\label{tab:dataset}
\end{table}

\subsubsection{Evaluation metrics}
Following previous multi-label image classification researches~\cite{ben2020asymmetric,cole2021multi}, mean Average Precision~(mAP) is used as the primary metric to evaluate model performance. 
We also report the overall F1-measure~(OF1) and per-category F1-measure~(CF1) with the positive threshold of 0.5. 
Precision~(P) and recall~(R) results are reported in Appendix B.

\subsubsection{Implementation details}
We use the Resnet50~\cite{he2016deep} pretrained on ImageNet~\cite{deng2009imagenet} as the backbone network for feature extraction. The input images are uniformly resized to $448\times448$. 
The optimizer is the Adam with a weight decay of 1e-4 and cycle learning rate schedule~\cite{smith2019super} is used with the max learning rate of 1e-4. Following~\cite{ben2020asymmetric}, we use the exponential moving average trick for higher baselines.

\subsubsection{Compared methods}
We firstly compare our methods with three classic baseline methods. 
Weak Assume Negatives~(WAN) and Label Smoothing~(LS)~\cite{cole2021multi} are selected as the stronger baselines than BCE in MLML. 
Secondly, we compare the Hill loss with loss re-weighting methods, Focal~\cite{lin2017focal} and the recently proposed ASL~\cite{ben2020asymmetric}. 
Thirdly, we choose the loss correction methods, including Pseudo Label~\cite{lee2013pseudo} and Regularized Online Label Estimation~(ROLE)~\cite{cole2021multi} to validate the effectiveness of the SPLC. 
For a fair comparison, we reproduce these methods using open-source codes from the authors under the same dataset configuration. We adopt the the hyper-parameters recommended in their papers.

\subsection{Comparisons with Existing Methods}
\subsubsection{Different missing ratios on COCO}
Table~\ref{tab:maintable} summarizes results of different methods on COCO under different missing ratios. As the increase of missing ratio, the model performance becomes worse and the efficacy of loss becomes significant. Though Weak Assume Negatives~(WAN) and Label Smoothing~(LS) are stronger baselines than BCE, the proposed Hill and SPLC show considerable superiority. 
Specifically, for the loss re-weighting methods, Hill significantly surpasses Focal and ASL by +3.49\% and +2.45\% respectively under the setting of 40\% labels left. 
For the loss correction methods, BCE + pseudo label is a two-stage method: we first train a model using BCE, afterwards use the trained model to correct the data and then retrain the model. ROLE requires additional space to store the label prediction matrix. In contrast, SPLC can correct missing labels in the training phase and does not need additional storage space. Therefore, SPLC improves the performance without any bells and whistles.

\begin{table*}[t]
\caption{Comparison results of mAP, CF1 and OF1 over other methods on COCO with different missing ratios.}
\begin{center}
\begin{tabular}{lccccccccc}
\hline
 &  \multicolumn{3}{c}{75\% labels left}&  
 \multicolumn{3}{c}{40\% labels left}& 
 \multicolumn{3}{c}{single label}
 \\
& 
\multicolumn{1}{c}{mAP} &
\multicolumn{1}{c}{CF1} &
\multicolumn{1}{c}{OF1} &
\multicolumn{1}{c}{mAP} &
\multicolumn{1}{c}{CF1} &
\multicolumn{1}{c}{OF1} &

\multicolumn{1}{c}{mAP} &

\multicolumn{1}{c}{CF1} &
\multicolumn{1}{c}{OF1} 

 \\
\hline

BCE~(full labels)  & 80.32 & 74.89 & 78.95 & - & - & - &- &- & - \\
BCE  & 76.81 & 67.73 & 71.07 & 70.49 & 45.81 & 40.42 & 68.57 & 43.78 & 37.72  \\
WAN~\cite{cole2021multi}  & 77.25 & 72.19 & 75.86 & 72.05 & 62.77 & 64.87 & 70.17 & 58.02 & 58.60  \\
BCE-LS~\cite{cole2021multi}  & 78.27 & 67.77 & 71.10 & 73.13 & 44.92 & 41.17 & 70.53 & 40.85 & 37.26  \\
\hline
Loss re-weighting:  \\
\hline
Focal~\cite{lin2017focal}  & 76.95 & 68.41 & 71.39 & 71.66 & 48.67 & 44.01 & 70.19 & 47.02 & 41.37  \\
ASL~\cite{ben2020asymmetric}  & 77.97 & 69.67 & 72.54 & 72.70 & 67.72 & 71.67 & 71.79 & 44.77 & 37.86  \\
\textbf{Hill~(Ours)}  & \textbf{78.84} & \textbf{73.57} & \textbf{77.31} & \textbf{75.15} & \textbf{68.56} & \textbf{73.08} & \textbf{73.17} & \textbf{65.47} & \textbf{69.51} \\
\hline
Loss correction: \\
\hline
BCE + pseudo label  & 77.05 & 68.18 & 71.47 & 71.46 & 52.17 & 46.48 & 69.77 & 48.91 & 42.99 \\
ROLE~\cite{cole2021multi}  & 78.43 &72.90  & \textbf{77.03} & 73.67 & 66.73 & 70.65 & 70.90 &57.59  & 61.16  \\
\textbf{\textbf{Focal margin + SPLC}~(Ours)}  & \textbf{78.44} & \textbf{73.18} & 76.55 & \textbf{75.69} & \textbf{67.91} & \textbf{73.33} & \textbf{73.18} & \textbf{61.55} & \textbf{67.35} \\
\hline


\end{tabular}
\end{center}
\label{tab:maintable}
\end{table*}

\begin{table}[t]
\caption{Comparison results of mAP on VOC-single label and NUS-single label datasets.}
\begin{center}
\scriptsize
\begin{tabular}{lcccccc}
\hline
 &  \multicolumn{1}{c}{VOC-single}&  
 \multicolumn{1}{c}{NUS-single}
\\
 \hline
BCE~(full labels)  & 89.04 & 60.63  \\
BCE  & 85.55 & 51.71 \\
WAN~\cite{cole2021multi}  & 87.03 & 52.89  \\
BCE-LS~\cite{cole2021multi}  & 87.23 & 52.47  \\
\hline
Loss re-weighting:        \\
\hline
Focal~\cite{lin2017focal}  & 86.83 & 53.58  \\
ASL~\cite{ben2020asymmetric}  & 87.33 & 53.92 \\
\textbf{Hill~(Ours)}  & \textbf{87.75} & \textbf{55.03} \\
\hline
Loss correction:        \\
\hline
BCE + pseudo label  & \textbf{89.67} & 51.79  \\
ROLE~\cite{cole2021multi}  & 89.00 & 50.61\\
\textbf{\textbf{Focal margin + SPLC}~(Ours)}  & 88.07 & \textbf{55.18}  \\
\hline

\end{tabular}
\end{center}
\label{tab:voc&nus}
\end{table}
  
\begin{table}[t]
\caption{Comparison results of mAP on Open Images-single label datasets.}
\begin{center}
\scriptsize
\begin{tabular}{ccccccc}
\hline
&
\multicolumn{1}{c}{BCE}&  
\multicolumn{1}{c}{Focal}&
\multicolumn{1}{c}{ASL}&
\multicolumn{1}{c}{Hill} &
\multicolumn{1}{c}{SPLC}
\\
\hline
Open Images  & 60.83 &62.14 &61.95  &  \textbf{62.71} & \textbf{62.86}  \\
\hline

\end{tabular}
\end{center}
\label{tab:openimages}
\vspace{-5mm}
\end{table}

\subsubsection{Single label on VOC, Nus-wide and Open Images}
The single positive label results on VOC and NUS are shown in Table~\ref{tab:voc&nus}. 
Hill and SPLC improve mAP performance remarkably on the NUS-single label dataset, but do not achieve obvious improvements on VOC. 
The reason may be that the VOC dataset only contains 20 labels, resulting in a relatively balanced ratio of positives and negatives. 
To further validate the effectiveness of proposed methods on extreme multi-label classification, we conduct experiments on Open Images with 567 labels as shown in Table~\ref{tab:openimages}.
It can be observed that both Hill and SPLC achieve improvements remarkably compared with common loss functions on Open Images, demonstrating that the proposed methods are robust in large-scale datasets and extreme classification cases.
\subsection{Analysis of Proposed Methods}

\begin{table}[t]
\caption{Ablation study of the Hill loss on the COCO-40\% labels left. Notations ``$+$'' and ``$-$'' represent the modification of positive and negative loss respectively on the basis of BCE.}
\begin{center}
\begin{tabular}{lcccc}
\hline
 &
mAP &
\multicolumn{1}{c}{OF1} & 
\multicolumn{1}{c}{CF1}    \\
\hline
 BCE  & 70.49 & 45.81 & 40.42 \\
\hline
Focal$^+$~\cite{lin2017focal}  & 70.87 & 49.42 & 44.63\\
\textbf{Focal margin$^+$~(Ours)} & \textbf{71.50} & \textbf{57.66} & \textbf{58.37} \\
 \hline
 WAN$^-$~\cite{cole2021multi}  &72.05 & 62.77 & 64.87\\
 Focal$^-$~\cite{lin2017focal} &72.10 & 56.43 & 54.04\\
 ASL$^-$~\cite{ridnik2021asymmetric}  & 72.70 & \textbf{67.72} & 71.67\\
 MSE$^-$ & 74.08 & 65.12 & 68.17\\
 \textbf{Hill$^-$~(Ours)}  & \textbf{74.98} & 66.56 & \textbf{71.87}\\
\hline

\textbf{Hill~(Ours)} & \textbf{75.15} & \textbf{68.56} & \textbf{73.08}\\
\hline
\end{tabular}
\end{center}
\label{tab:ablation}
\vspace{-3mm}
\end{table}

\subsubsection{Ablation study of Hill loss}
Since Hill loss is composed of positive and negative loss, we conduct the ablation study on the basis of BCE to verify the effectiveness of positive and negative parts in Hill loss as shown in Table~\ref{tab:ablation}. 
For the positive part, the hard-mining method~(Focal$^+$) brings a slight gain for BCE, whereas Focal margin$^+$ brings a relatively significant improvement. This shows semi-hard mining is more appropriate for positives than hard-mining in MLML. 
For the negative part, we first down-weight the whole negatives with a weight parameter termed WAN$^-$, which provides a stronger baseline than BCE with solid improvements on different metrics.
Second, we compare different re-weighting losses. Hill$^-$ surpasses Focal$^-$ and ASL$^-$ by above +2\% mAP score. The main reason is that though ASL$^-$ seeks to attenuate the effect of hard negatives, the gradient analysis in Fig.~\ref{img:posloss} illustrates that ASL$^-$ still puts too much focus on these possibly false negatives compared with Hill$^-$.
Third, the superiority of Hill$^-$ over MSE$^-$ also demonstrates the necessity of re-weighting.
The performance is further improved via the integration of Focal margin$^+$ and Hill$^-$. 

\begin{figure*}[t]
\flushleft
	\subfigure{
    	\begin{minipage}[h]{0.25\linewidth}
    	\centering
    	\includegraphics[width=1.7in]{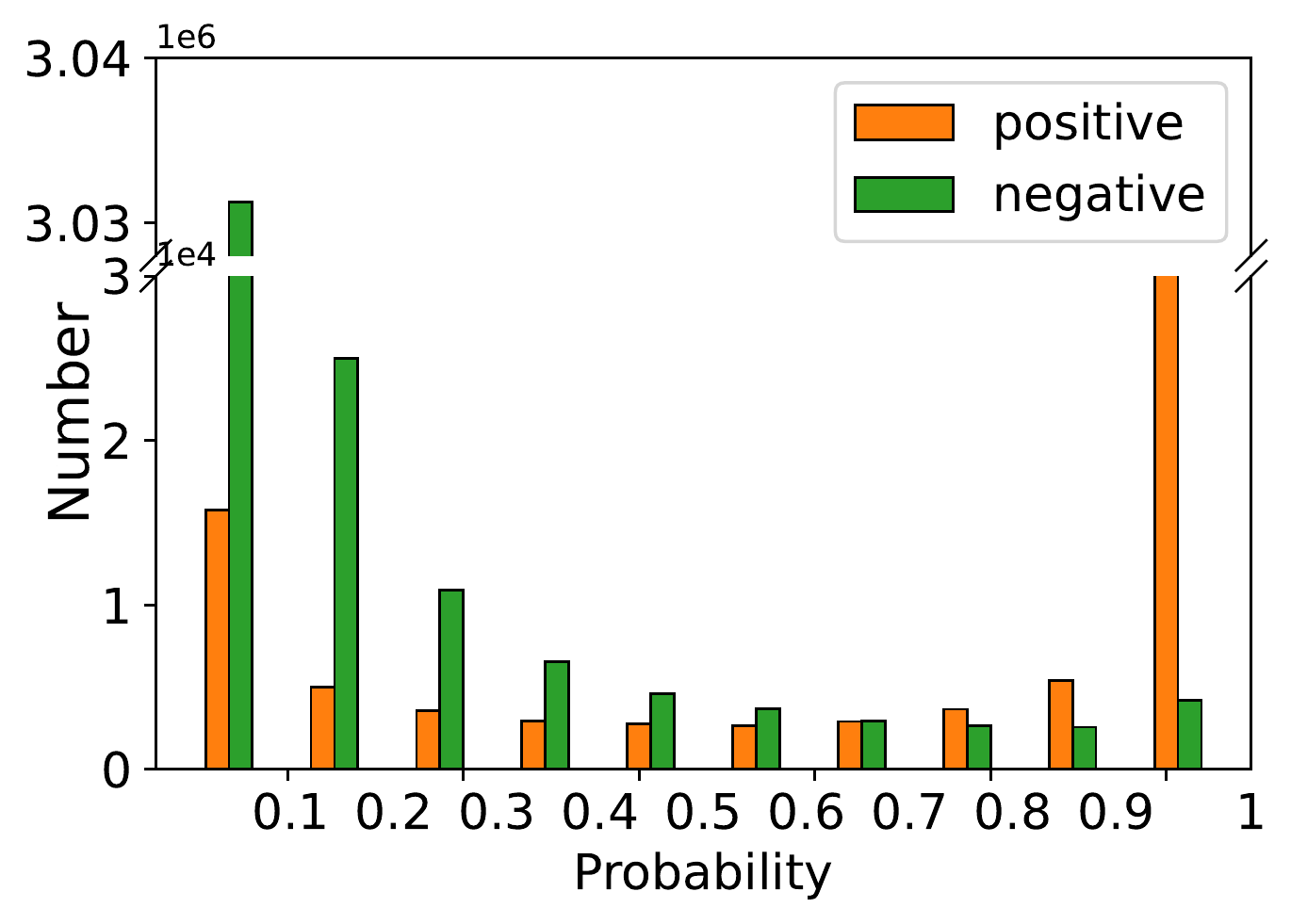}
    	\scriptsize (a)~BCE-full labels
    	\includegraphics[width=1.7in]{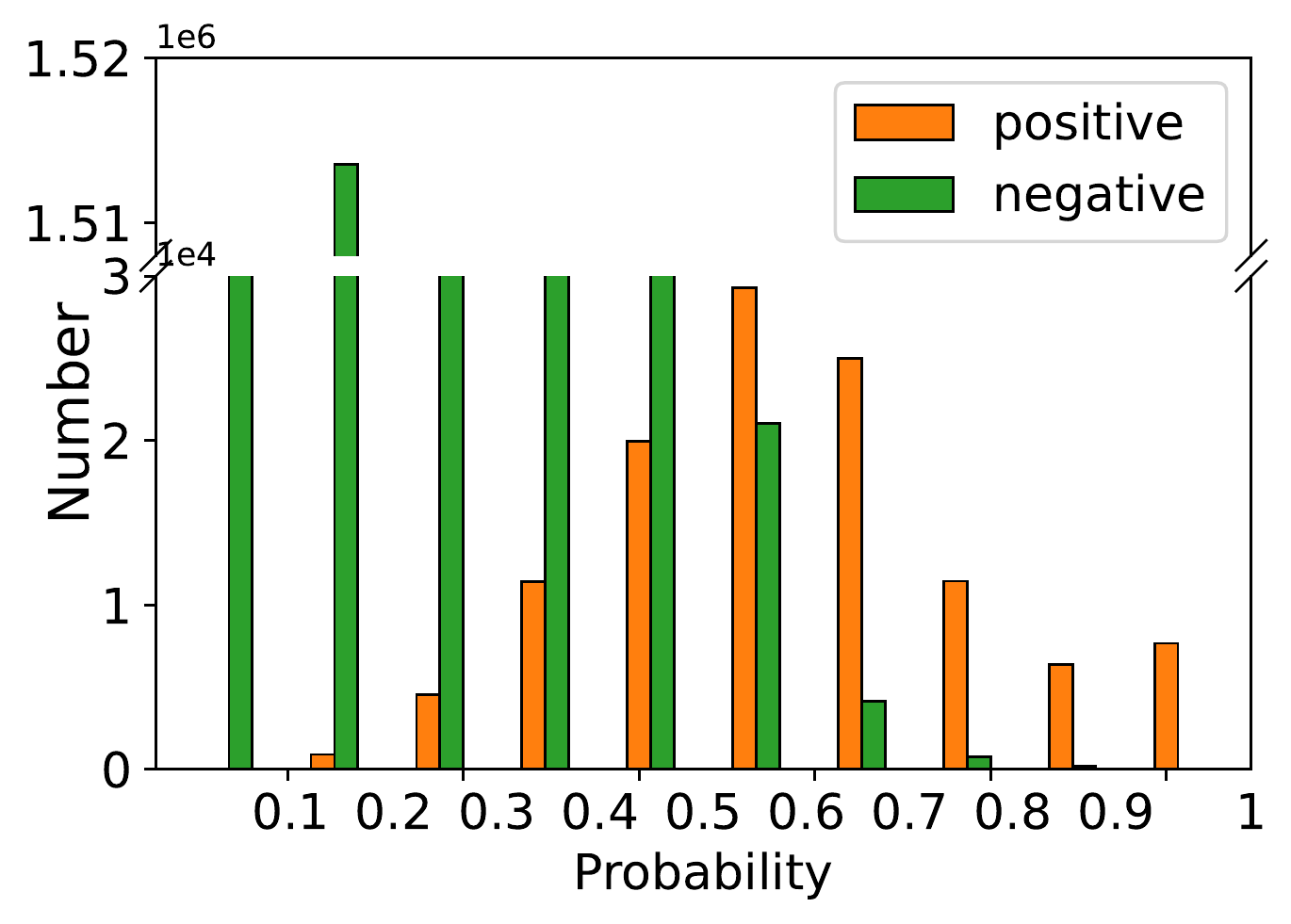}
    	\scriptsize (e)~ASL
	    \end{minipage}%
	}%
	\subfigure{
    	\begin{minipage}[h]{0.25\linewidth}
    	\centering
    	\includegraphics[width=1.7in]{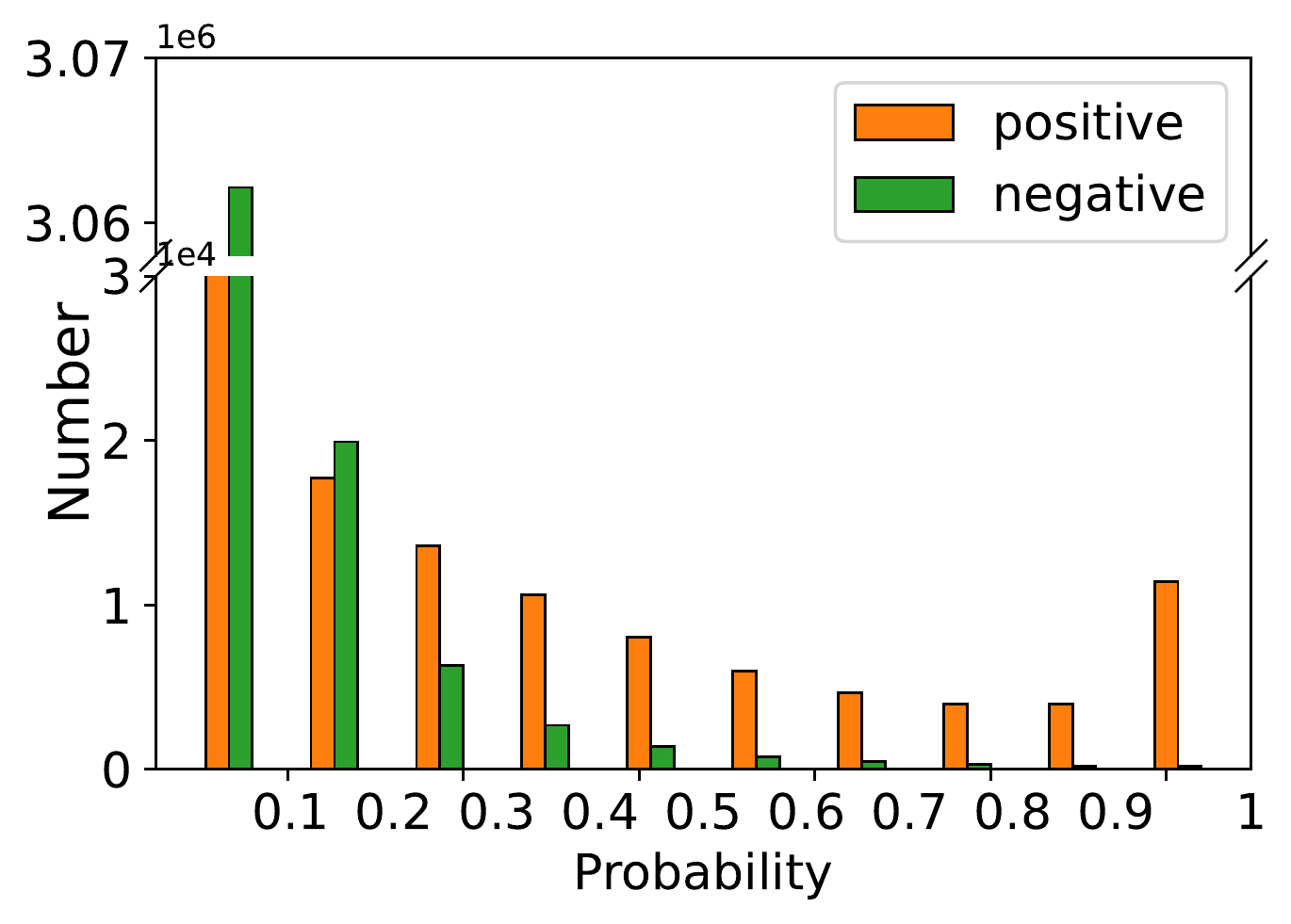}
    	\scriptsize (b)~BCE
    	\includegraphics[width=1.7in]{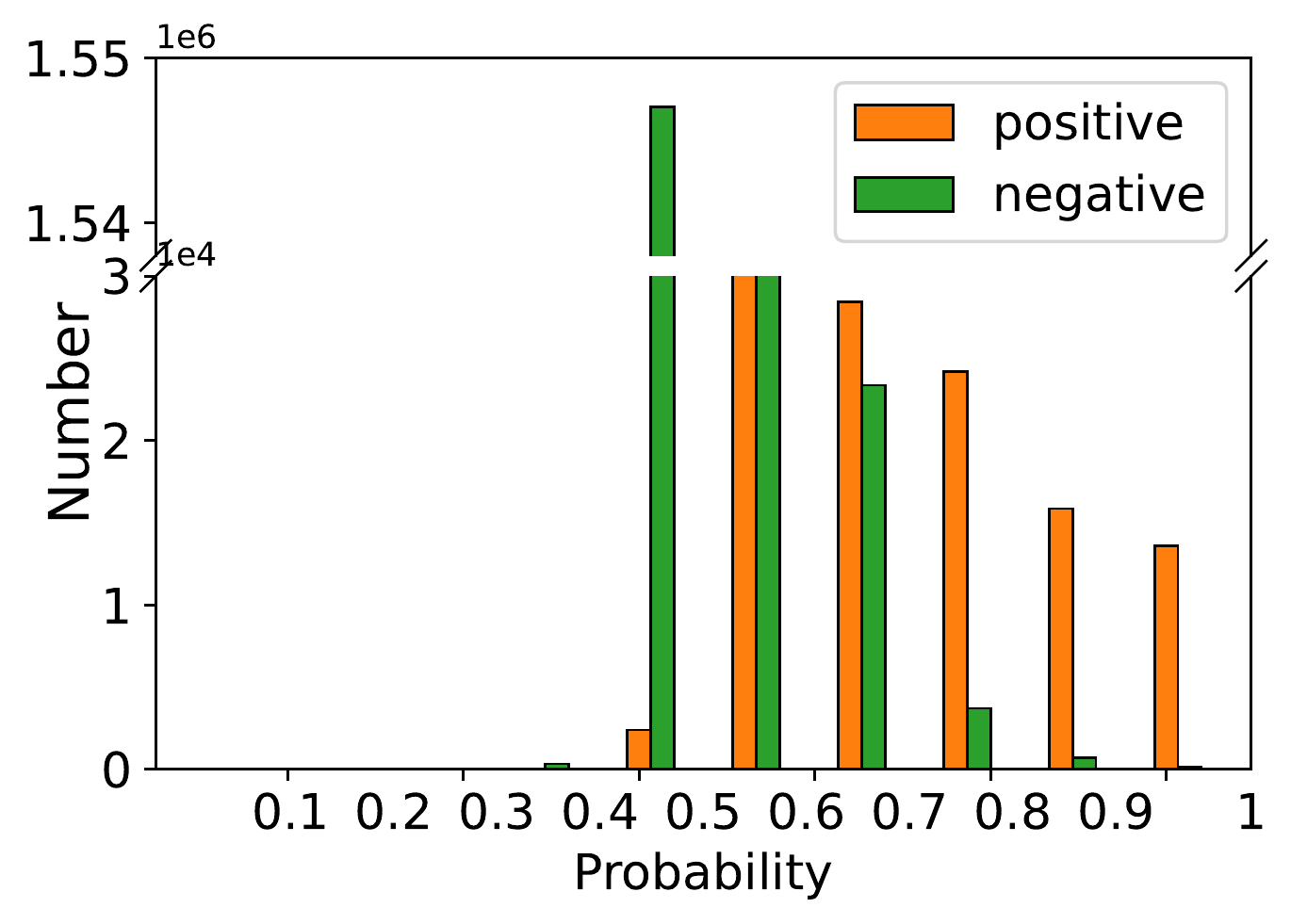}
	    \scriptsize (f)~ROLE
    	\end{minipage}%
	}%
	\subfigure{
    	\begin{minipage}[h]{0.25\linewidth}
    	\centering
    	\includegraphics[width=1.7in]{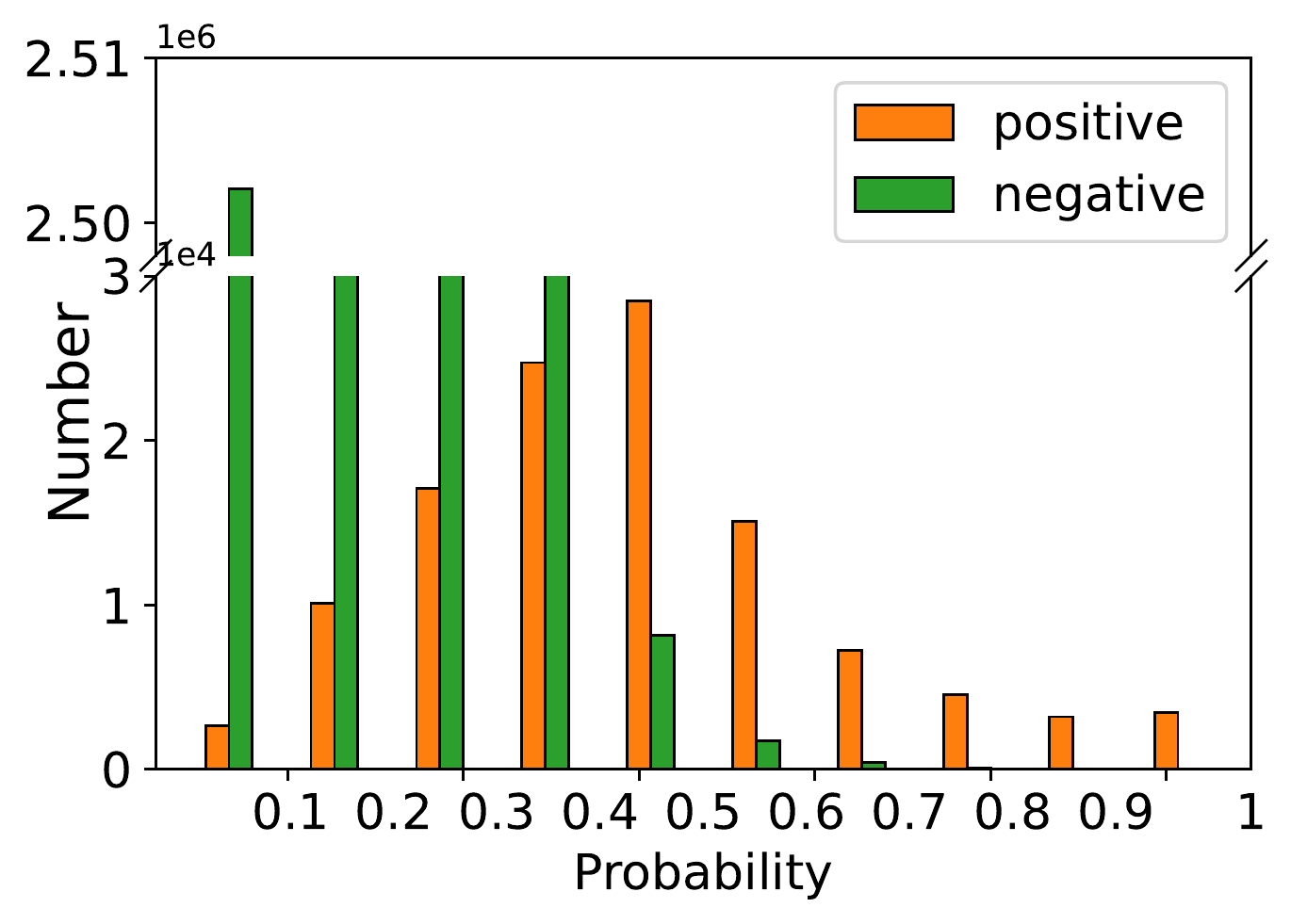}
    	\scriptsize (c)~Focal
    	\includegraphics[width=1.7in]{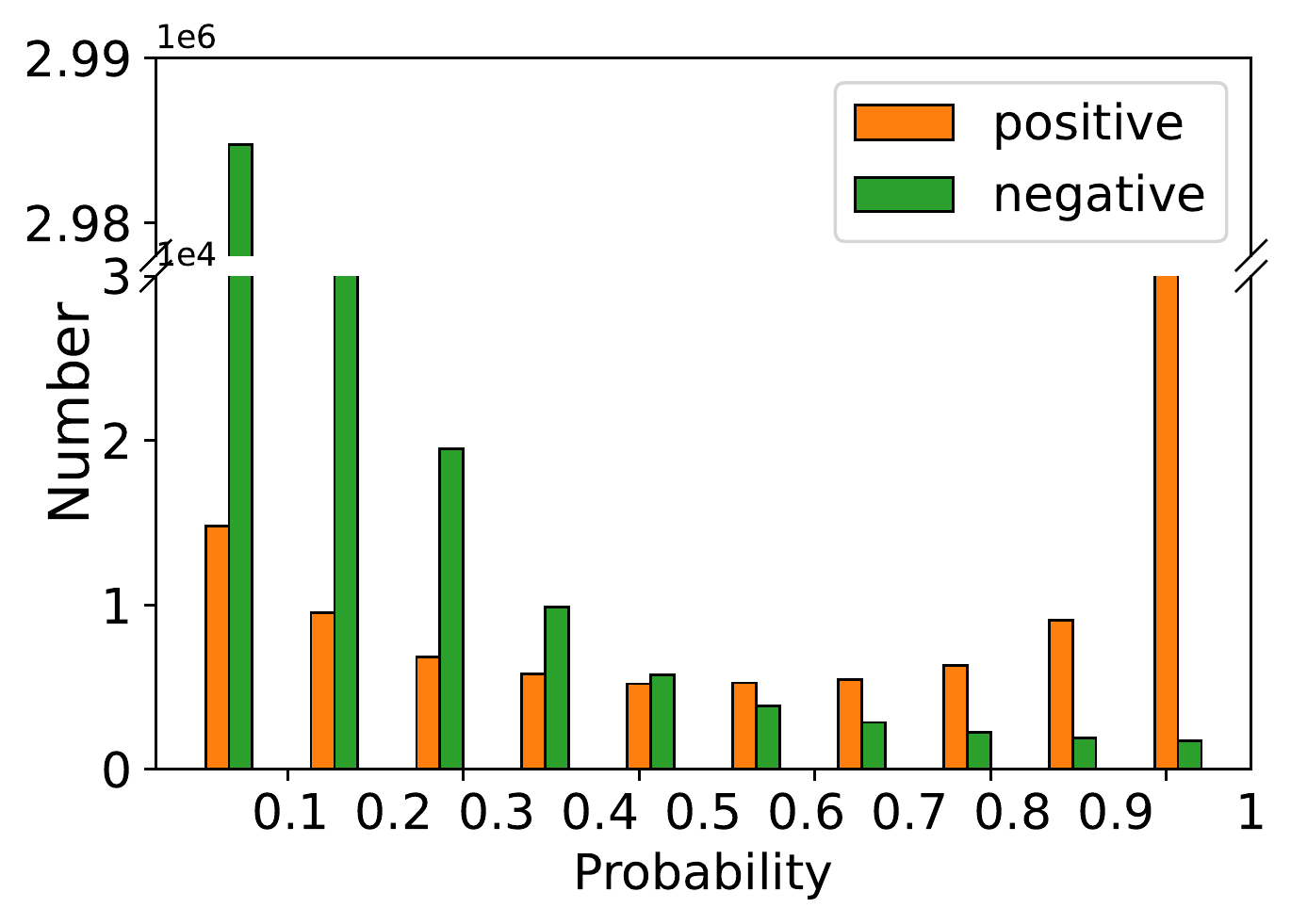}
    	\scriptsize (g)~Hill~(Ours)
    	\end{minipage}%
	}%
	\subfigure{
    	\begin{minipage}[h]{0.25\linewidth}
    	\centering
    	\includegraphics[width=1.7in]{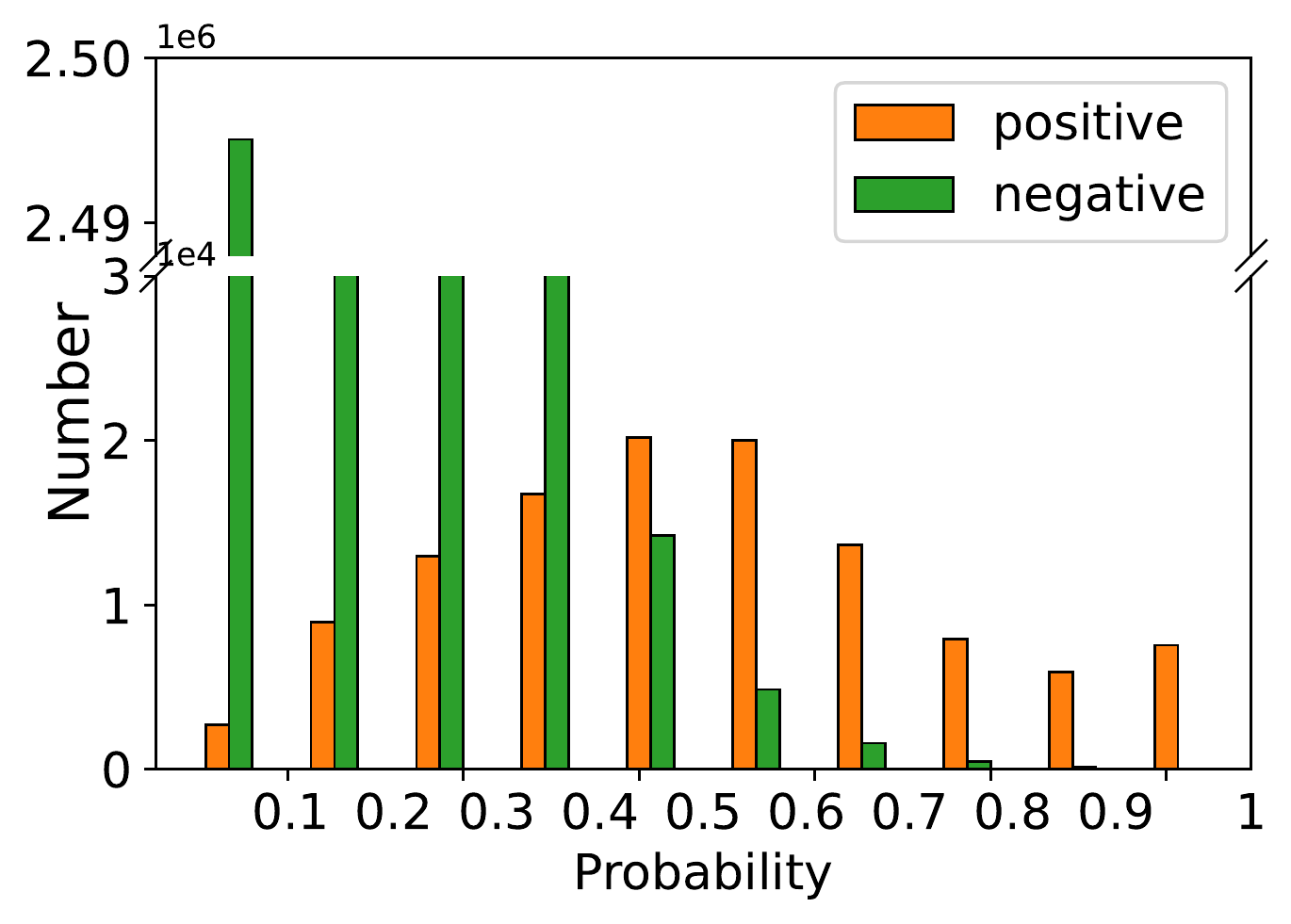}
    	\scriptsize (d)~Focal margin~(Ours)
    	\includegraphics[width=1.7in]{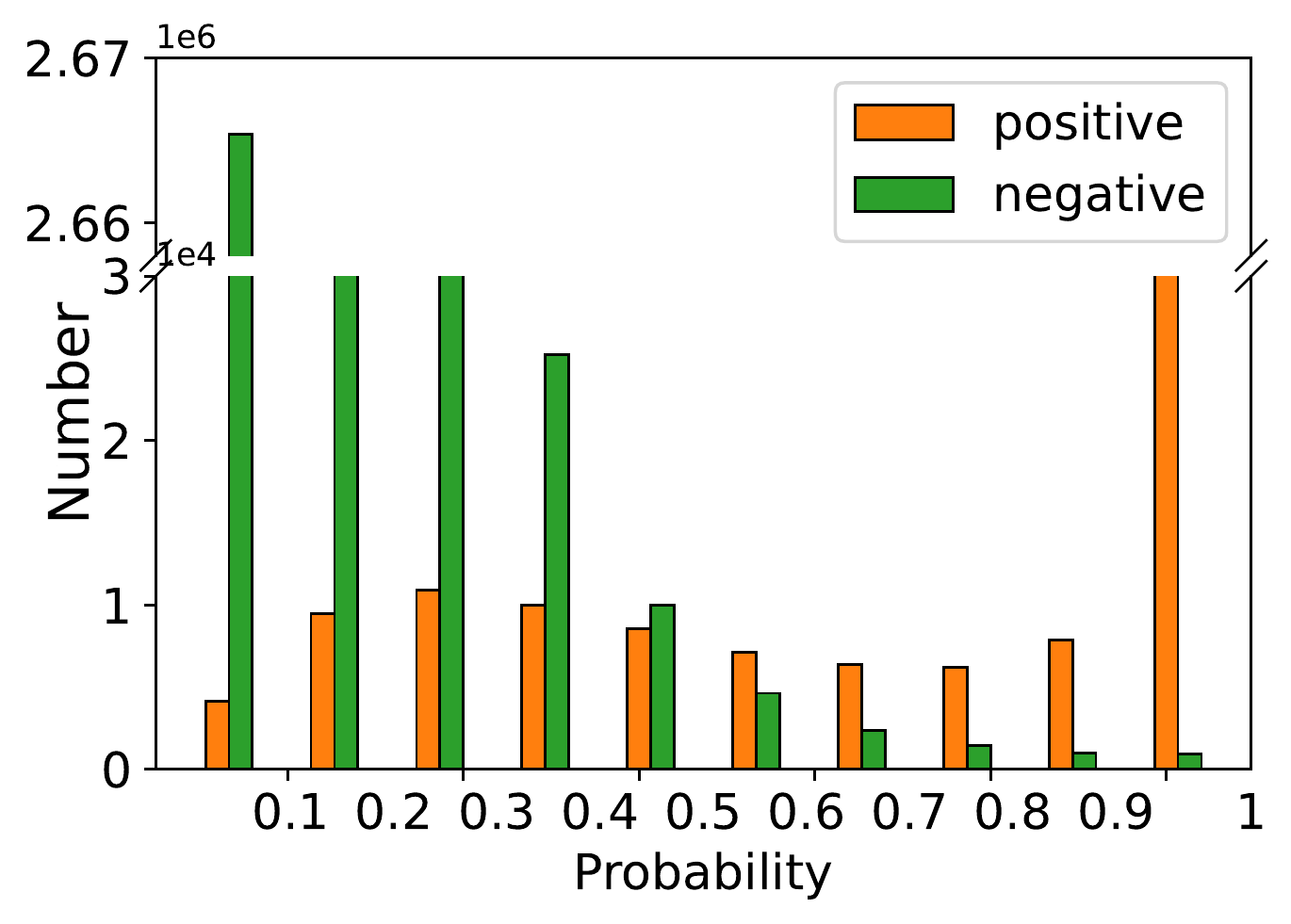}
    	\scriptsize (h)~SPLC~(Ours)
    	\end{minipage}%
	}%
	\centering
	\caption{Probability distributions on the COCO test set of different loss functions. Note that only (a) is obtained from a model trained with full labels on the COCO dataset, and the rest of the figures are obtained from models trained via different methods on the COCO-40\% labels left.}
\label{img:dis}
\end{figure*}

\begin{table}[t]
\caption{Quantitative results of SPLC with different loss functions on the COCO-40\% labels left.}
\begin{center}
\scriptsize
\begin{tabular}{lccc}
\hline
Method
 &  \multicolumn{1}{c}{mAP}&  
 \multicolumn{1}{c}{CF1}&
  \multicolumn{1}{c}{OF1}
\\
\hline
BCE  & 70.49 & 45.81 & 40.42  \\
BCE + pseudo label  & 71.46 & 52.17 & 46.48  \\
BCE + \textbf{SPLC}  & \textbf{73.63} & \textbf{65.56} &\textbf{72.14} \\
\hline
Focal  & 71.66 & 48.67 &44.01 \\
Focal + pseudo label  & 72.67 & 52.43 & 47.68  \\
Focal + \textbf{SPLC}  & \textbf{73.83} & \textbf{57.95} &\textbf{56.04} \\
\hline
ASL & 72.70 & 67.72  &71.67 \\
ASL + pseudo label  & 73.75 & 68.73 & 72.06  \\
ASL + \textbf{SPLC} & \textbf{74.23} & \textbf{69.23} & \textbf{72.74} \\
\hline
Focal margin & 72.26 & 60.79  &61.58 \\
Focal margin + pseudo label  & 74.10 & 65.53 & 66.17  \\
\textbf{Focal margin + SPLC} & \textbf{75.69} & \textbf{67.91} &\textbf{73.33} \\
\hline

\end{tabular}
\end{center}
\label{tab:SPLC ablation}
\vspace{-5mm}
\end{table}

\subsubsection{SPLC boosts existing losses}
Table~\ref{tab:SPLC ablation} shows that both pseudo label and SPLC methods can complement existing losses and lead to consistent improvements, while SPLC performs better than the pseudo label method in both training efficiency and accuracy.
SPLC corrects the missing labels dynamically along with the training, while pseudo label method needs an extra training process to predict the labels.

\begin{figure}[t]
  \centering
  \includegraphics[width=0.40\textwidth]{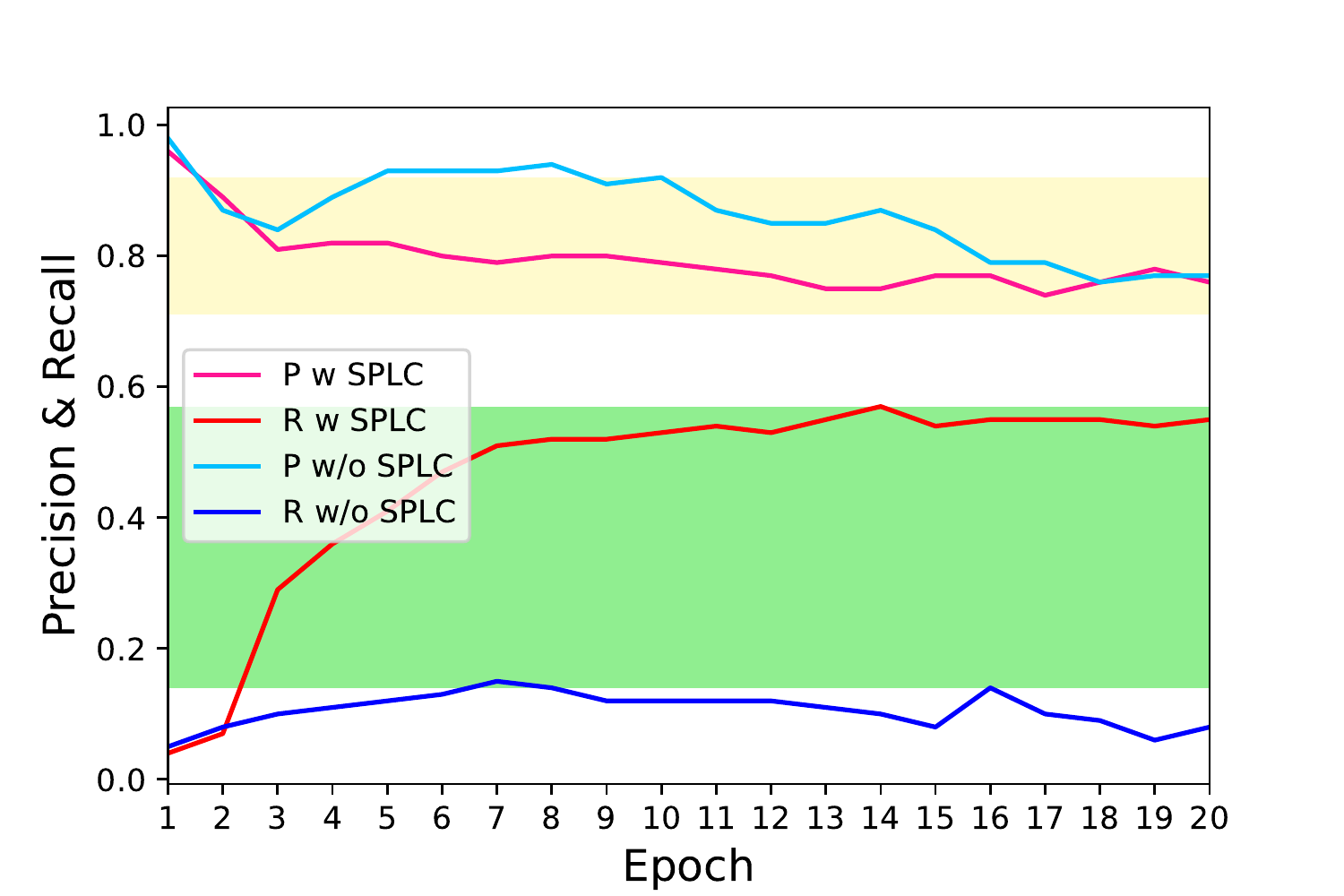}
  \caption{The precision and recall curves of missing labels during training on the COCO training set of 40\% labels left with and without SPLC. The yellow and green rectangles refer to the precision and recall of a well-trained model used for pseudo labeling with the thresholds between 0.4 and 0.6.}
  \label{img:SPLC_pr}
\end{figure}

Fig.~\ref{img:SPLC_pr} is drawn to explain why SPLC achieves better accuracy. 
It can be observed that during the training process, SPLC gradually recalls more missing labels and correct them, guaranteeing the precision meanwhile.  
By comparison, it is hard for the pseudo label method to make a good compromise between precision and recall, since the model is trained by noisy datasets and is not robust to the missing labels.

\begin{figure*}[t]
\flushleft
  \centering
  \includegraphics[width=1.0\textwidth]{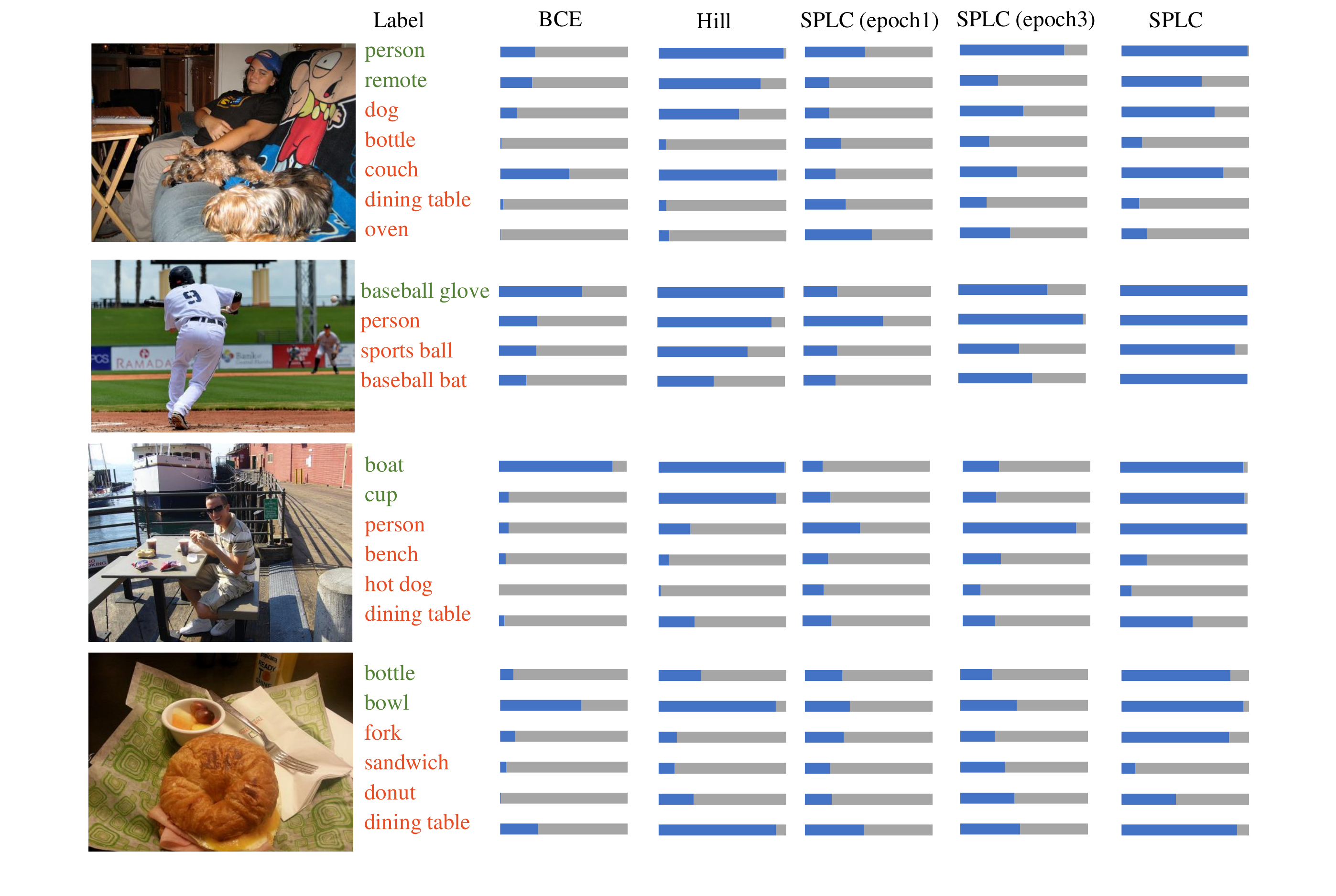}
	\centering
	\caption{Example results of our proposed methods and BCE from the COCO test set. Green and red color mean positive and missing labels respectively. Generally, the model trained with BCE tends to overfit to the missing labels, as their predicted probabilities are quite low.
	By comparison, models trained by Hill and SPLC predict higher probabilities for missing labels, implying that they are more robust against missing labels.
	We also show the results from the first and third epoch of SPLC to demonstrate the feasibility and rationality of loss correction at the early stage.
	More examples are shown in Appendix B.}
\label{img:example_vis}
\end{figure*}

Moreover, Focal margin performs best when combined with SPLC, which obtains a remarkable mAP gain up to 3.43\%. The reason is that the Focal margin highlights semi-hard positives as shown in Fig.~\ref{img:posloss}, leading to more missing labels being corrected than the Focal loss.

\subsubsection{Probability distribution analysis}
\label{sec:discuss}
Fig.~\ref{img:dis} illustrates the probability distributions of different methods. 
As shown in Fig.~\ref{img:dis}~(g) and (h), our proposed methods outperform existing works obviously, as positives and negatives are well differentiated and the probability distribution are similar to that of BCE trained on the COCO full labels dataset. 
Specifically, decreasing the weighting of hard negatives and correcting noisy labels mitigate the effects of mislabeled samples. Besides, paying more emphasis on semi-hard positives is conducive to the learning of positive samples.
More concretely,
From Fig.~\ref{img:dis}~(a) and (b), it can be observed that many positives are incorrectly identified under the impact of missing labels. 
Comparing Fig.~\ref{img:dis}~(c) and (d), we observe that more positives are recalled by Focal margin loss than Focal loss, which indicates that semi-hard mining is conducive on multi-label image recognition. 
As for ASL~(Fig.~\ref{img:dis}~(f)), although it demonstrates the effectiveness of down-weighting negatives, it only down-weights extremely hard negatives~($p>0.8$), leaving semi-hard negative samples~($p\in[0.5,0.8]$) large weights. 
As a result, too large weights for hard negatives and too small weights for easy ones make insufficient distinction between positives and negatives. 
Despite higher mAP obtained by ROLE than baseline methods, the distribution of ROLE suffers the problem of insufficient discrimination especially for the negatives, as shown in Fig.~\ref{img:dis}~(f).
Even more, Role requires additional memory to store the label matrix for online estimation, which is not negligible for large-scale datasets.

Furthermore, we give examples of the predicted probabilities from different methods in Fig.~\ref{img:example_vis}. Hill and SPLC present better robustness against missing labels, as they allocate relatively higher probability values to missing labels than BCE.

\subsection{Additional Discussions and Experiments}
\subsubsection{Impact of the hyper-parameters}
Despite the introduction of extra hyper-parameters in Hill and SPLC compared with BCE, we can empirically select them as described in the method part. Here we give more detailed analysis and answer some common questions in practical use.
In principle, we did not invest too much efforts in finding the best hyper-parameters for each dataset. 
The optimal parameters searched on the COCO-40\% labels left are shared for all other datasets.

\textit{How to set $m$ of Focal margin?}
The margin~$m$ of Focal margin controls the importance of different positives. A small $m$ focuses more on hard positives,
and the focus shift to semi-hard positives as $m$ increases. Based on our empirical evaluation in Table~\ref{tab:hyper-parameter}, we recommend $m\in[0.5,1.0]$ and set $m=1$ in all the experiments.

\textit{How to determine the certain time of early stage for SPLC?}
SPLC is insensitive to the determination of the certain time of early stage.
Experiments show that a selection from the 2-nd to 5-th epoch has little effect~(mAP within 0.2) on performance shown in Table~\ref{tab:hyper-parameter}.
The reason is that the model recalls only a small part of FNs~(but with high precision) at the early stage.
Therefore, we correct the labels after the first epoch for all the experiments of SPLC.

\textit{What if FNs are identified incorrectly for SPLC?} 
In fact, it cannot be guaranteed that all identified FNs are correct, even in well-studied single-label tasks.
However, our methods would perform well as long as most identified FNs are correct, as demonstrated by the extensive experiments. 
The error rate of false differentiation in training set is below 5\% by setting a high threshold~(above 0.6) in Fig.~\ref{img:observation}. 
The underlying reason behind this is the distinctive features in MLML as described in Sec.~\ref{sec_observation}.

\textit{How to set the threshold of SPLC to differentiate TNs and FNs?}
The threshold $\tau$ controls the trade-off between precision and recall of FNs as shown in Fig.~\ref{img:SPLC_pr}.
$\tau$ is a relatively sensitive parameters as shown in Table~\ref{tab:hyper-parameter}. 
A higher~(stricter) threshold means less FNs are recalled but the precision is higher.
We set the threshold $\tau$ to a fixed high value of 0.6 to alleviate the incorrect identification of FNs.
The reason why $\tau$ is greater than the commonly used 0.5 is that the Focal margin focuses more on positives and recalls more missing labels compared with the BCE.
It is predictable that setting different thresholds for different classes and training stages, instead of a fixed one, will bring better accuracy but more parameters to be tuned. 
Thus, we leave the study of adaptive and efficient threshold for the future work.

\textit{Does the Hill loss adjust the ratio of positive and negative part by the hyper-parameters, similar to Focal loss and ASL?}
Hill loss has a fixed formulation for positive and negative part but outperforms Focal loss and ASL. Therefore, no hyper-parameter is needed to balance positives and negatives.


\begin{table}[t]
\begin{center}
\caption{MAP results of different margins in Focal margin, different epochs and thresholds in SPLC on the COCO-40\% labels left.}
\label{tab:hyper-parameter}
\begin{tabular}{c|ccccc}
\hline
\multicolumn{6}{c}{\textbf{Different margins~$(m)$}}\\
\hline
$m$ &
\multicolumn{1}{c}{0}&  
\multicolumn{1}{c}{0.5}&
\multicolumn{1}{c}{1.0}&
\multicolumn{1}{c}{1.5} &
\multicolumn{1}{c}{2.0} \\
mAP  & 73.57 &75.01 &\textbf{75.15}  &  74.81 & 74.53  \\
\hline
\multicolumn{6}{c}{\textbf{Different epochs}}\\
\hline
epoch  & 1 &2 &3  &  4 & 5  \\
mAP  &75.69  &75.63 &75.60  &75.70  &75.71   \\
\hline
\multicolumn{6}{c}{\textbf{Different thresholds~$(\tau)$}}\\
\hline
threshold  & 0.5 &0.55 &0.6  &  0.65 &0.7  \\
mAP  &55.84  &72.41 & \textbf{75.69} & 75.46 & 74.29  \\
\hline
\end{tabular}
\end{center}
\end{table}

\subsubsection{More experiments with different backbones}
In Table~\ref{tab:backbones}, we test the applicability of Hill and SPLC for different
backbones, by comparing the different loss functions on three representative architectures: MobilenetV3-Large~\cite{howard2019searching}, Resnet101~\cite{he2016deep} and Swin-transformer-Small~\cite{liu2021swin}. 
The consistent improvements brought by Hill and SPLC demonstrate that our methods are robust to different backbones.

\begin{table}[t]
\caption{Comparison results of mAP with different backbones on COCO-40\% labels left.}
\begin{center}
\scriptsize
\begin{tabular}{lcccccc}
\hline
 &  \multicolumn{1}{c}{MobilenetV3}&  
 \multicolumn{1}{c}{Resnet101} & 
  \multicolumn{1}{c}{Swin-transformer}
\\
\hline
BCE~(full labels)  &  76.32 &81.74 &83.31  \\
BCE  & 68.45 &71.92 &76.74 \\
WAN~\cite{cole2021multi}  & 69.27 & 73.58 &77.20  \\
BCE-LS~\cite{cole2021multi}  &68.73  &74.41 &76.88  \\
Focal~\cite{lin2017focal}  &69.46  &73.58 &76.78 \\
ASL~\cite{ben2020asymmetric}  & 69.65 &75.09 &76.70  \\
\textbf{Hill~(Ours)}  & \textbf{71.03} & \textbf{76.53} & \textbf{78.96}\\
\textbf{\textbf{Focal margin + SPLC}~(Ours)}  & \textbf{71.10} & \textbf{77.20} &\textbf{78.19}  \\
\hline

\end{tabular}
\end{center}
\label{tab:backbones}
\vspace{-2mm}
\end{table}

\section{Conclusion} 

In this paper, we designed simple and robust loss functions for multi-label learning with missing labels without sacrificing training and inference efficiency. 
First, we presented an observation that partial missing labels can be distinguished from negatives in the early stage of training with a high precision. 
Then, two novel approaches are proposed to alleviate the effect of missing labels by down-weighting and correcting potential missing labels, respectively. 
Besides, we proposed to perform semi-hard mining on positives, which further promoted the performance.
Extensive experiments on several widely used benchmarks have demonstrated the superiority of the proposed methods.
Our methods substantially simplify the existing pipeline of MLML.
We hope the new state-of-the-arts established by our methods can serve as a starting point for future research on MLML.

\bibliographystyle{IEEEtran}
\bibliography{reference}
\cleardoublepage
\appendices
\section{}
\subsection{Gradient Analysis of ASL}
\begin{figure}[ht]
	\subfigure{
    	\begin{minipage}[t]{0.5\linewidth}
    	\centering
    	\includegraphics[width=1.7in]{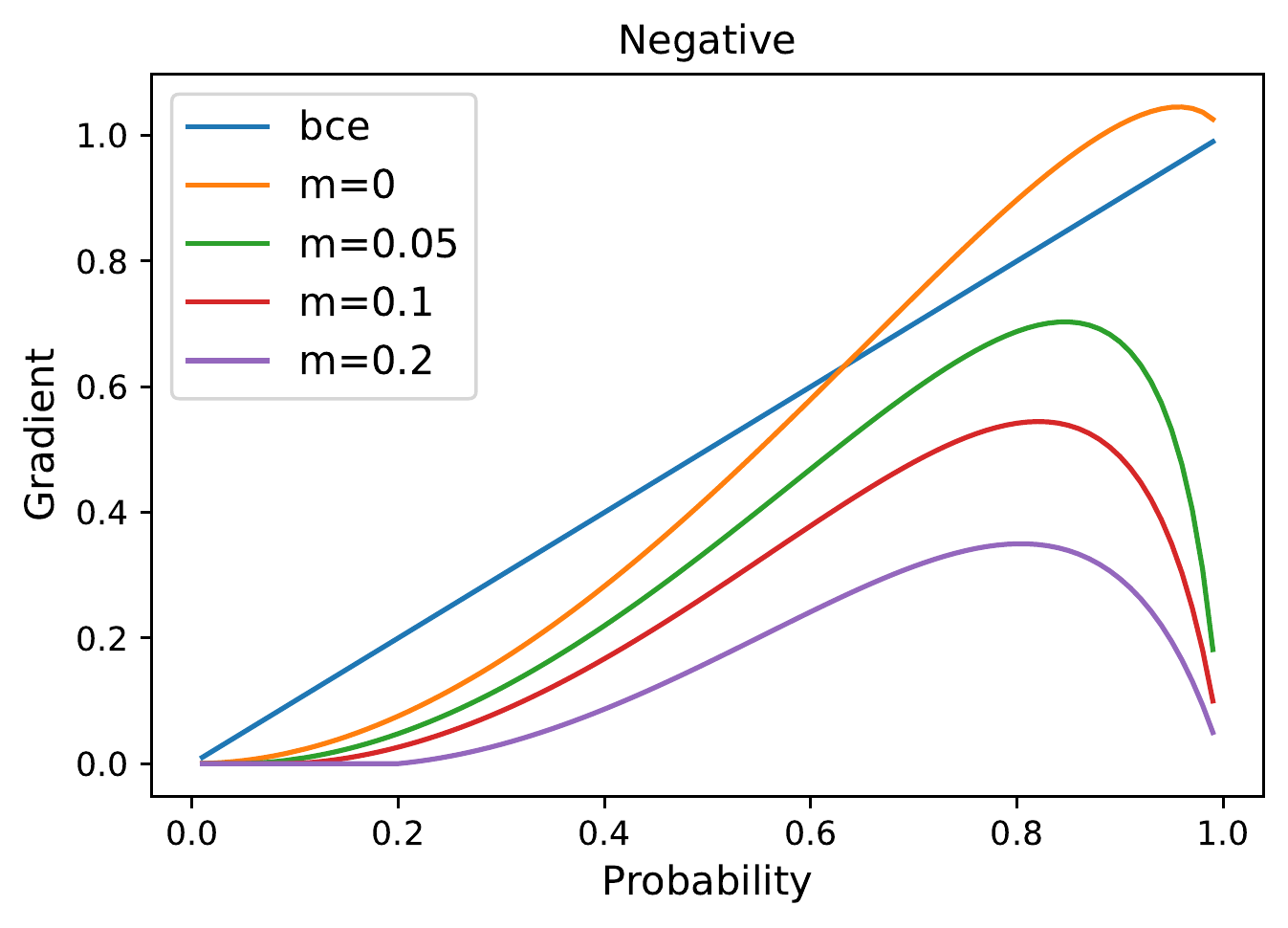}
    	\scriptsize (a)~ASL-$\gamma=1$
    	\centering
    	\includegraphics[width=1.7in]{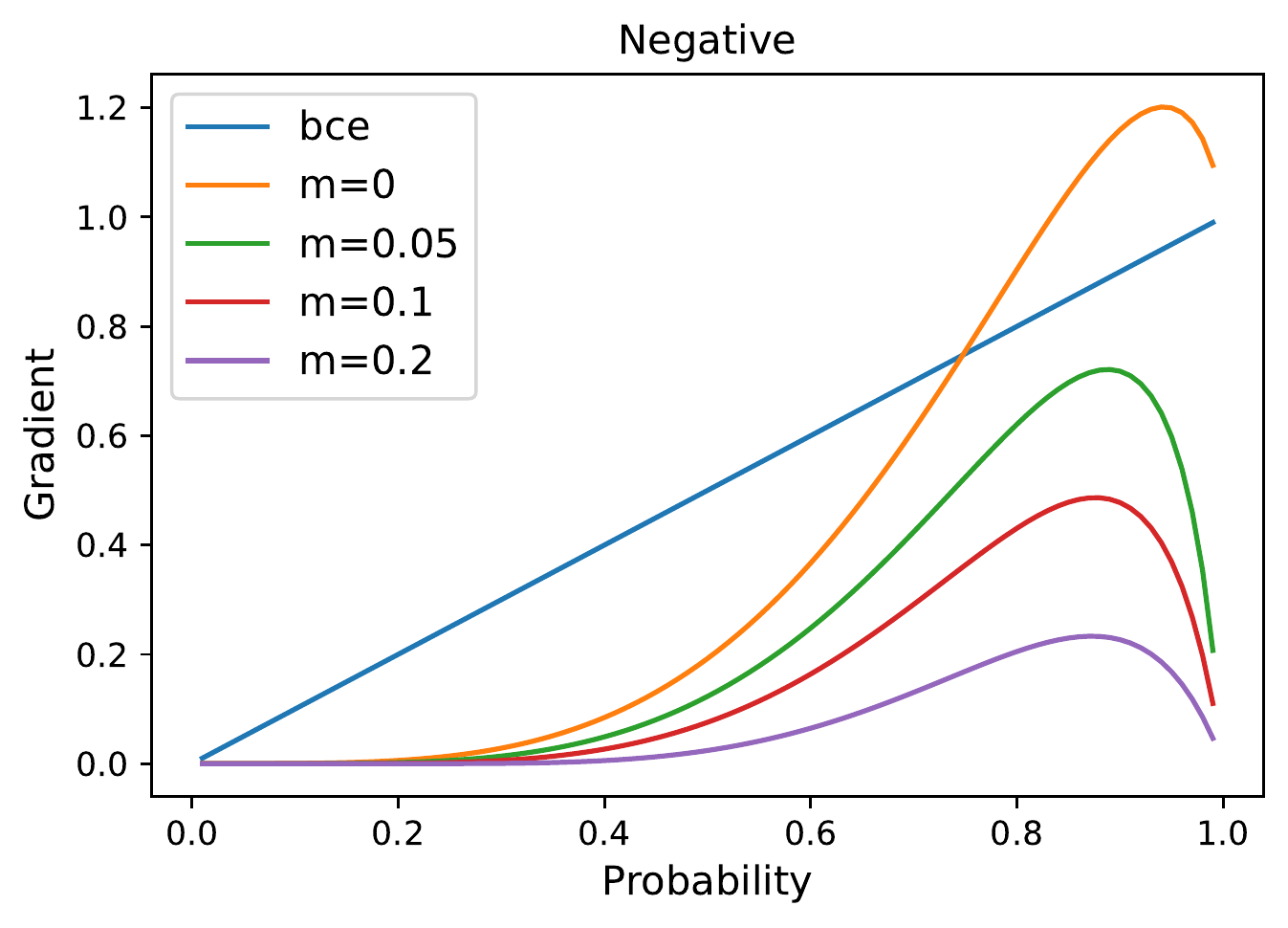}
    	\scriptsize (c)~ASL-$\gamma=3$
	    \end{minipage}%
	}%
	\subfigure{
    	\begin{minipage}[t]{0.5\linewidth}
    	\centering
    	\includegraphics[width=1.7in]{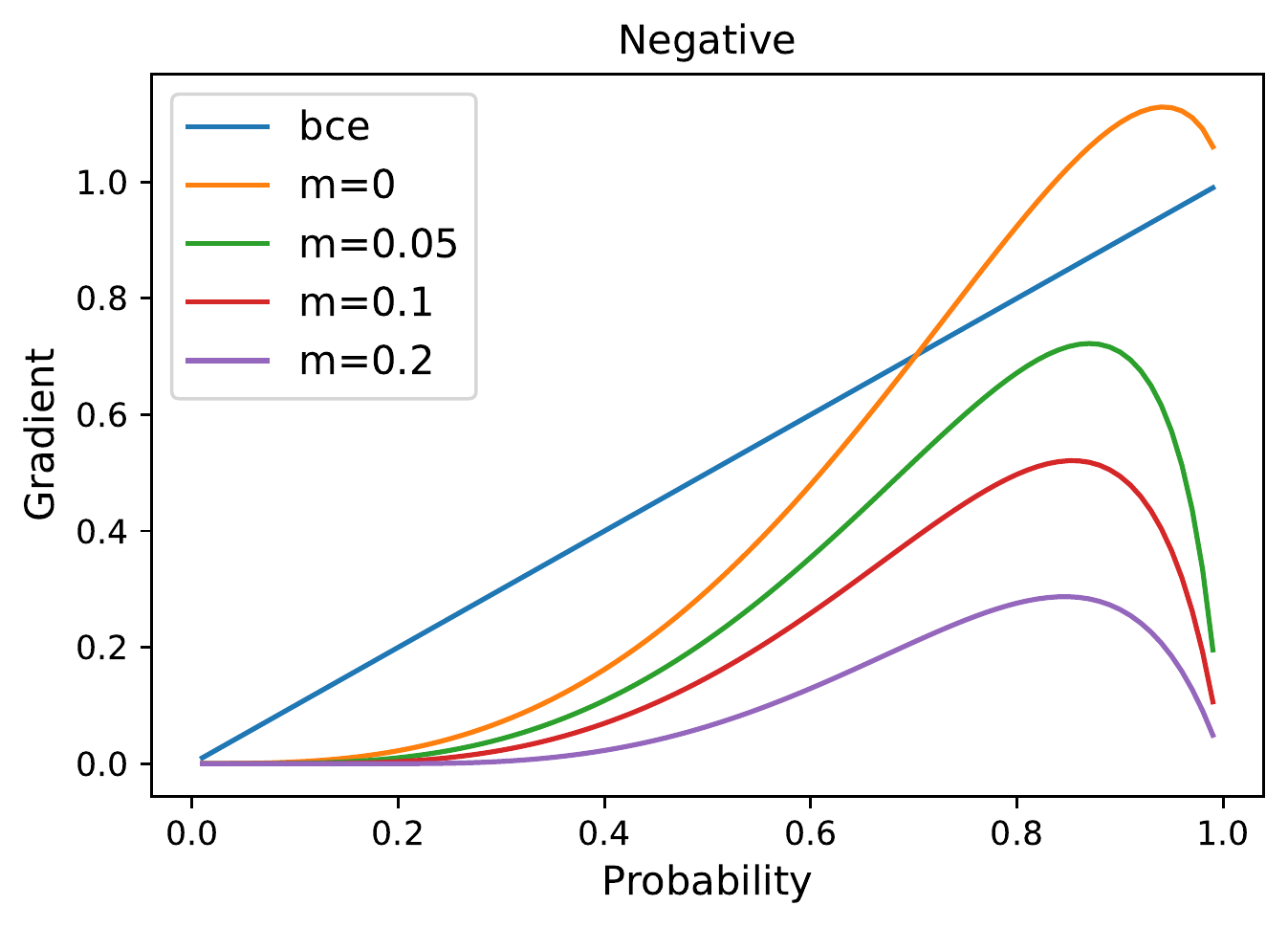}
    	\scriptsize (b)~ASL-$\gamma=2$
    	\centering
    	\includegraphics[width=1.7in]{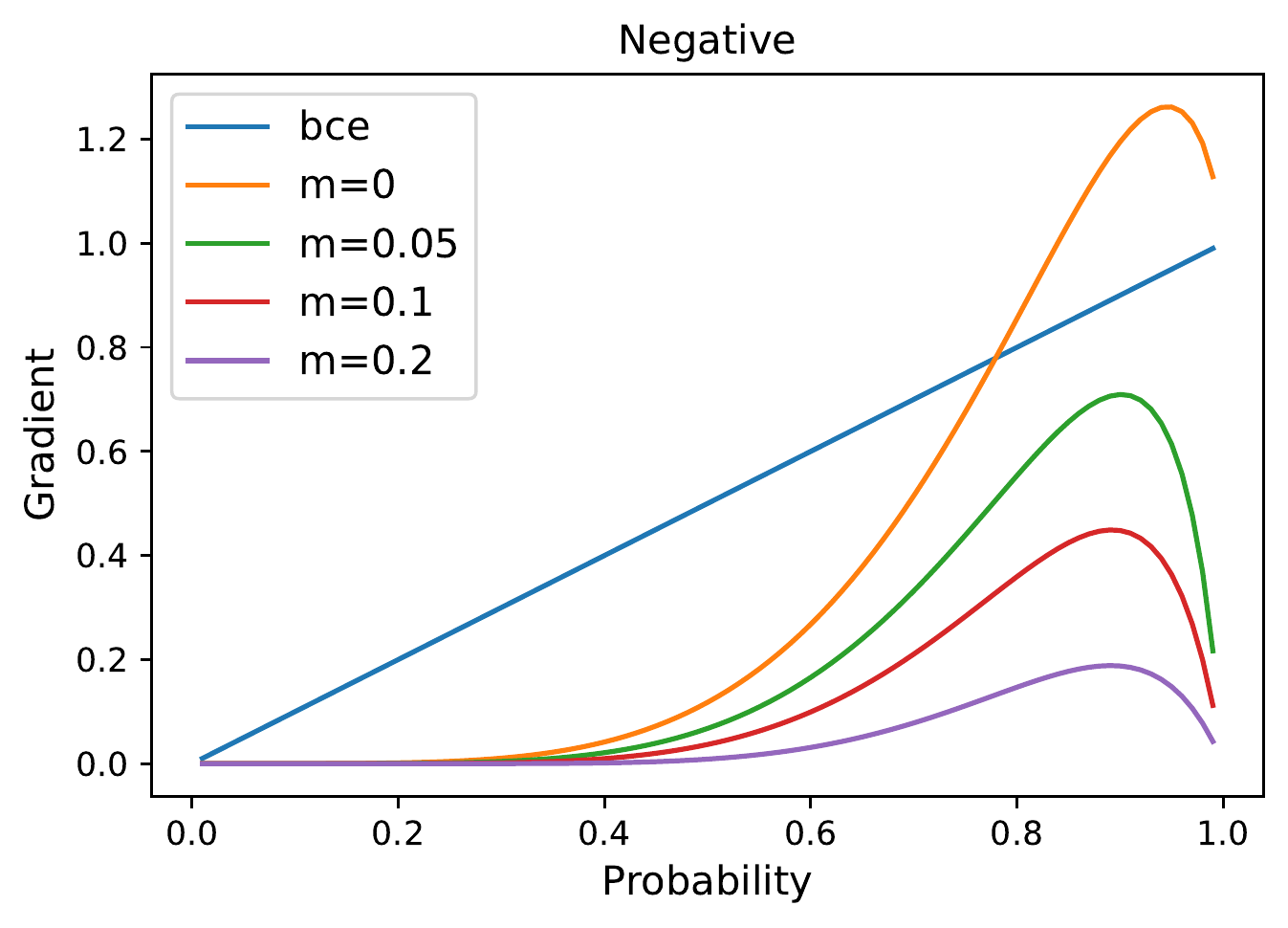}
    	\scriptsize (d)~ASL-$\gamma=4$
    	\end{minipage}%
	}%
	\centering
	\caption{Gradient analysis of ASL loss with different $\gamma$ and $m$. ASL only down-weights extremely hard negatives~($p>0.9$), but leaves semi-hard negative samples~($p\in[0.5,0.8]$) large weights.}
    \label{img:app_asl}
\end{figure}

The negative gradients of ASL~\cite{ben2020asymmetric} with different $\gamma$ and $m$ are shown in Fig.~\ref{img:app_asl}. 
Despite attempts at different hyper-parameters, the turning point is limited as $p\in[0.8.0.9]$. 
Given that, ASL still focuses too much on most possibly missing labels samples~(false negatives with $p>0.5$).

\subsection{Additional Evaluation Metrics and Examples}
Following previous multi-label image classification researches~\cite{ben2020asymmetric,cole2021multi},mean Average Precision~(mAP) is adopted as the primary metric to evaluate model performance. 
In Table~\ref{tab:app_coco0.5}, Table~\ref{tab:app_coco0.2}, Table~\ref{tab:app_cocosingle}, Table~\ref{tab:app_vocsingle}, Table~\ref{tab:app_nussingle} and Table~\ref{tab:app_openimagesingle} respectively, we report additional evaluation results including OF1 CF1, OP, OR, CP and CR on different datasets. 
The positive threshold is set to a commonly used value of 0.5 for all methods, except ROLE~\cite{cole2021multi}. As shown in Fig.~\ref{img:dis}~(f), the ROLE has a threshold offset~(the lowest prediction value $p_l$ is generally around $0.3$ rather than $0$). Therefore, the positive threshold is set to $\tau=(1+p_l)/2$ for a relatively fair comparison.

The results show that our proposed Hill loss and SPLC outperform existing methods. Precision~(P) and Recall~(R) are trade-off metrics, which are sensitive to the selected threshold. Thus, we should take a comprehensive look at the two metrics, instead of focusing each single metric.

\begin{figure*}[ht]
\flushleft
  \centering
  \includegraphics[width=1.0\textwidth]{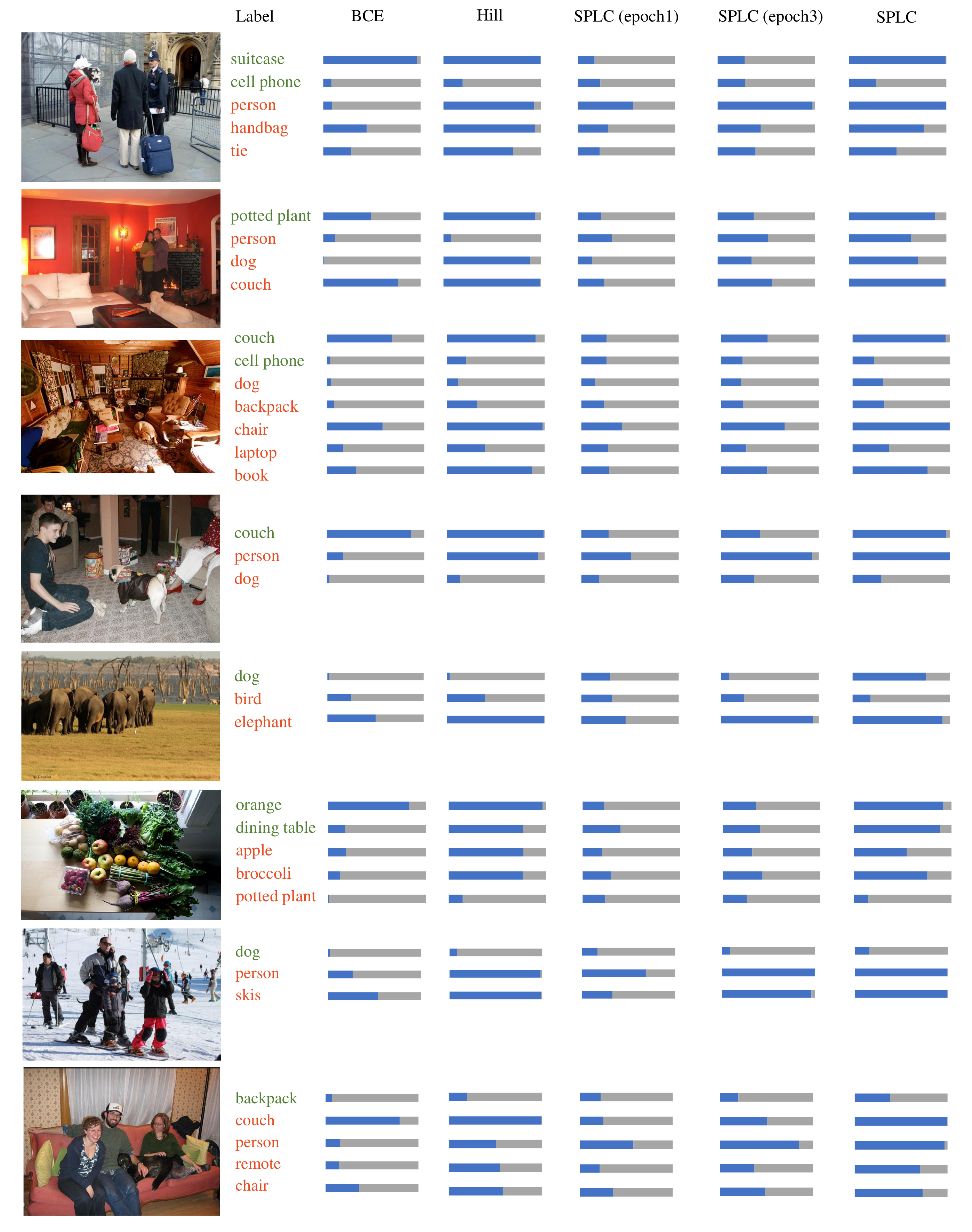}
	\centering
   \vspace{-1em}
	\caption{More example results of our proposed methods and BCE from the COCO test set. Green and red color mean positive and missing labels respectively.}
\label{img:example_vis_app}
\end{figure*}

More example results of our proposed methods and BCE from the COCO test set are provided in Fig.~\ref{img:example_vis_app}.

\begin{table}[ht]
\caption{Comparison results of various metrics over other methods on the COCO-40\% labels left. }
\begin{center}
\setlength{\tabcolsep}{1.2mm}{
\begin{tabular}{lcccccccccccccccccccccccc}
\hline
& 
\multicolumn{1}{c}{mAP} &
\multicolumn{1}{c}{CP} &
\multicolumn{1}{c}{CR} &
\multicolumn{1}{c}{CF1} &
\multicolumn{1}{c}{OP} &
\multicolumn{1}{c}{OR} &
\multicolumn{1}{c}{OF1}&
 \\
\hline
BCE~(full labels)  & \underline{80.3} &\underline{80.8} & \underline{70.3} & \underline{74.9} &\underline{84.3} &\underline{74.2} & \underline{78.9}  \\
BCE  & 70.5 &89.2 &34.4  & 45.8 &94.1 & 25.7  & 40.4\\
WAN~\cite{cole2021multi}  & 72.1 &81.0 &53.2 & 62.8 &86.3 &52.0   & 64.9 \\
BCE-LS~\cite{cole2021multi}  & 73.1 & 92.9 &33.5  & 44.9 &96.4 &26.2  & 41.2 \\
\hline
Sample-reweighting:  \\
\hline
Focal~\cite{lin2017focal}  & 71.7 &\textbf{88.9} &37.0 & 48.7 &\textbf{93.7} &28.8 & 44.0 \\
ASL~\cite{ben2020asymmetric}  & 72.7 &71.4 &\textbf{65.6} & 67.7 &75.3 &\textbf{68.4} & 71.7 \\
\textbf{Hill~(Ours)}  & \textbf{75.2} &80.4 &61.4& \textbf{68.6} & 85.5 &63.8 &\textbf{73.1}  \\
\hline
Loss correction: \\
\hline
BCE + pseudo label  & 71.5 &\textbf{88.0} &40.8  & 52.2 &\textbf{92.4} & 31.1 & 46.5 \\
ROLE~\cite{cole2021multi}  & 73.7 &78.9 &60.0 &66.7 &83.6 &61.2 &70.7 \\
\textbf{\textbf{SPLC}~(Ours)}  & \textbf{75.7} & 81.6 & \textbf{60.7} &  \textbf{67.9} &87.7 & \textbf{63.0}  & \textbf{73.3} \\

\hline
\end{tabular}
\label{tab:app_coco0.2}
}
\end{center}
\end{table}

\begin{table}[ht]
\caption{Comparison results of various metrics over other methods on the COCO-75\% labels left. }
\begin{center}
\setlength{\tabcolsep}{1.2mm}{
\begin{tabular}{lcccccccccccccccccccccccc}
\hline
& 
\multicolumn{1}{c}{mAP} &
\multicolumn{1}{c}{CP} &
\multicolumn{1}{c}{CR} &
\multicolumn{1}{c}{CF1} &
\multicolumn{1}{c}{OP} &
\multicolumn{1}{c}{OR} &
\multicolumn{1}{c}{OF1} &
 \\
\hline
BCE~(full labels)  & \underline{80.3} &\underline{80.8} & \underline{70.3} & \underline{74.9} &\underline{84.3} &\underline{74.2} & \underline{78.9}  \\
BCE  & 76.8 &85.1 &58.1  & 67.7 &90.1 & 58.7  & 71.1\\
WAN~\cite{cole2021multi}  & 77.3 &75.0 &70.2 & 72.2 &78.2 &73.7   & 75.9 \\
BCE-LS~\cite{cole2021multi}  & 78.3 & 88.2 &57.2  & 67.8 &92.1 &57.9  & 71.1 \\
\hline
Loss re-weighting:  \\
\hline
Focal~\cite{lin2017focal}  & 77.0 &\textbf{83.8} &59.4 & 68.4 &\textbf{88.6} &59.8 & 71.4 \\
ASL~\cite{ben2020asymmetric}  & 77.9 &63.1 &\textbf{78.8} & 69.7 &64.9 &\textbf{82.2} & 72.5 \\
\textbf{Hill~(Ours)}  & \textbf{78.8} &73.6 &74.4& \textbf{73.6} &76.4 &78.3 & \textbf{77.3} \\
\hline
Loss correction: \\
\hline
BCE + pseudo label  & 77.1 &\textbf{84.9} &58.7  & 68.2 &\textbf{89.7} & 59.4 & 71.5 \\
ROLE~\cite{cole2021multi}  & \textbf{78.4} &77.4 &70.0 &72.9 &80.9 &73.5 & \textbf{77.0} \\
\textbf{\textbf{SPLC}~(Ours)}  & \textbf{78.4} & 72.6 & \textbf{75.1} &  \textbf{73.2} &74.0 & \textbf{79.3}  & 76.6 \\

\hline

\end{tabular}
\label{tab:app_coco0.5}
}
\end{center}
\end{table}

\begin{table}[!htbp]
\caption{Comparison results of various metrics over other methods on the COCO-single label. }
\begin{center}
\setlength{\tabcolsep}{1.2mm}{
\begin{tabular}{lcccccccccccccccccccccccc}
\hline
& 
\multicolumn{1}{c}{mAP} &
\multicolumn{1}{c}{CP} &
\multicolumn{1}{c}{CR} &
\multicolumn{1}{c}{CF1} &
\multicolumn{1}{c}{OP} &
\multicolumn{1}{c}{OR} &
\multicolumn{1}{c}{OF1} &
 \\
\hline
BCE~(full labels)  & \underline{80.3} &\underline{80.8} & \underline{70.3} & \underline{74.9} &\underline{84.3} &\underline{74.2} & \underline{78.9}  \\
BCE  & 68.6 &88.6 &33.0  & 43.8 &93.9 & 23.6  & 37.7\\
WAN~\cite{cole2021multi}  & 70.2 &82.4 &47.9 & 58.0 &88.2 &43.9   & 58.6 \\
BCE-LS~\cite{cole2021multi}  & 70.5 & 89.3 &30.8  & 40.9 &96.5 &23.1  & 37.3 \\
\hline
Loss re-weighting:  \\
\hline
Focal~\cite{lin2017focal}  & 70.2 &88.2 &36.0 & 47.0 & 93.4 & 41.4 &26.6 \\
ASL~\cite{ben2020asymmetric}  & 71.8 &\textbf{90.4} &33.4 & 44.8 &\textbf{94.7} &23.7 & 37.9 \\
\textbf{Hill~(Ours)}  & \textbf{73.2} &79.7 &\textbf{58.0}& \textbf{65.5} &85.3 &\textbf{58.7} & \textbf{69.5} \\
\hline
Label correction: \\
\hline
BCE + pseudo label  & 69.8 &\textbf{87.2} &38.3  & 48.9 &\textbf{92.5} & 28.0 & 43.0 \\
ROLE~\cite{cole2021multi}  & 70.9 &83.3 &49.1 &57.6 &88.3 &46.8 & 61.2 \\
\textbf{\textbf{SPLC}~(Ours)}  & \textbf{73.2} & 83.8 & \textbf{53.1} &  \textbf{61.6} &90.1 & \textbf{53.8}  & \textbf{67.4} \\

\hline

\end{tabular}
\label{tab:app_cocosingle}
}
\end{center}
\end{table}

\begin{table*}[ht]
\caption{Comparison results of various metrics over other methods on the VOC-single label. }
\begin{center}
\setlength{\tabcolsep}{1.2mm}{
\begin{tabular}{lcccccccccccccccccccccccc}
\hline
& 
\multicolumn{1}{c}{mAP} &
\multicolumn{1}{c}{CP} &
\multicolumn{1}{c}{CR} &
\multicolumn{1}{c}{CF1} &
\multicolumn{1}{c}{OP} &
\multicolumn{1}{c}{OR} &
\multicolumn{1}{c}{OF1} &

 \\
\hline
BCE~(full labels)  & \underline{89.0} &\underline{82.9} &\underline{84.1}  & \underline{83.3} &\underline{86.4} & \underline{84.7}  & \underline{85.6}\\
BCE  & 85.6 &87.8 &71.6  & 77.4 &91.1 & 68.7  & 78.4\\
WAN~\cite{cole2021multi}  & 87.0 &83.9 &78.4 & 80.4 &87.5 &77.7   & 82.3 \\
BCE-LS~\cite{cole2021multi}  & 87.2 & 92.0 &67.7  & 75.4 &94.6 &65.9  & 77.7 \\
\hline
Sample-reweighting:  \\
\hline
Focal~\cite{lin2017focal}  & 86.8 &\textbf{87.2} &73.6 & 78.2 & \textbf{90.6} & 70.3 &79.1 \\
ASL~\cite{ben2020asymmetric}  & 87.3 &68.4 &\textbf{87.2} & 75.9 &72.8 &\textbf{86.9} & 79.2 \\
\textbf{Hill~(Ours)}  & \textbf{87.8} &85.3 &78.7& \textbf{81.1} &88.5 &79.6 & \textbf{83.8} \\
\hline
Label correction: \\
\hline
BCE + pseudo label  & \textbf{89.7} &86.5 &\textbf{81.8}  & \textbf{84.0} &89.1 & \textbf{83.5} & \textbf{86.2} \\
ROLE~\cite{cole2021multi}  & 89.0 &87.6 &78.2 &81.5 &90.6 &79.0 & 84.4 \\
\textbf{\textbf{SPLC}~(Ours)}  & 88.1 &\textbf{88.7}  & 75.0 &  80.2 &\textbf{91.9} & 75.8  & 83.0 \\

\hline

\end{tabular}
\label{tab:app_vocsingle}
}
\end{center}
\end{table*}

\begin{table*}[t]
\caption{Comparison results of various metrics over other methods on the NUS-single label dataset. }
\begin{center}
\setlength{\tabcolsep}{1.2mm}{
\begin{tabular}{lcccccccccccccccccccccccc}
\hline
& 
\multicolumn{1}{c}{mAP} &
\multicolumn{1}{c}{CP} &
\multicolumn{1}{c}{CR} &
\multicolumn{1}{c}{CF1} &
\multicolumn{1}{c}{OP} &
\multicolumn{1}{c}{OR} &
\multicolumn{1}{c}{OF1} &
 \\
\hline
BCE~(full labels)  & \underline{60.6} &\underline{62.7} &\underline{58.0}  &\underline{59.1} &\underline{73.8} & \underline{73.4}  &\underline{73.6}\\
BCE  & 51.7 &67.8 &23.2  & 30.9 &84.0 & 20.7  & 33.3\\
WAN~\cite{cole2021multi}  & 52.9 &61.5 &41.2 & 47.2 &75.1 &46.7   & 57.6 \\
BCE-LS~\cite{cole2021multi}  & 52.5 & 66.9 &19.8  & 27.0 &85.6 &21.0  &33.7 \\
\hline
Sample-reweighting:  \\
\hline
Focal~\cite{lin2017focal}  & 53.6 &\textbf{69.1} &26.2 & 34.2 & \textbf{83.4} & 22.1 &35.0 \\
ASL~\cite{ben2020asymmetric}  & 53.9 &53.5 &55.0 & 53.2 &67.7 &67.3 & 67.5 \\
\textbf{Hill~(Ours)}  & \textbf{55.0} &54.7 &\textbf{56.3}& \textbf{54.1} &68.5 &\textbf{68.7} & \textbf{68.6} \\
\hline
Label correction: \\
\hline
BCE + pseudo label  & 51.8 &\textbf{66.8} &24.8  & 32.5 &\textbf{83.6} & 22.2 & 35.1 \\
ROLE~\cite{cole2021multi}  & 50.6 &56.3 &31.7 &37.4 &75.7 &50.8 &60.8  \\
\textbf{\textbf{SPLC}~(Ours)}  & \textbf{55.2} &56.6  & \textbf{54.0} &  \textbf{52.4} & 68.4& \textbf{73.0}  & \textbf{70.6} \\

\hline

\end{tabular}
\label{tab:app_nussingle}
}
\end{center}
\end{table*}

\begin{table*}[ht]
\caption{Comparison results of various metrics over other methods on the Open Images-single label dataset. }
\begin{center}
\setlength{\tabcolsep}{1.2mm}{
\begin{tabular}{lcccccccccccccccccccccccc}
\hline
& 
\multicolumn{1}{c}{mAP} &
\multicolumn{1}{c}{CP} &
\multicolumn{1}{c}{CR} &
\multicolumn{1}{c}{CF1} &
\multicolumn{1}{c}{OP} &
\multicolumn{1}{c}{OR} &
\multicolumn{1}{c}{OF1} &

\\
\hline
BCE  &60.8  &\textbf{72.9} &30.6  &38.2  &\textbf{85.9} &14.5   &24.9 \\
Focal~\cite{lin2017focal}   &62.1  &72.3 &32.3  &39.8  &85.6 &15.2   &25.8  \\
ASL~\cite{ben2020asymmetric}   &62.0  &60.1 &55.5  &53.4  &69.2 &38.3   &49.3  \\
\textbf{Hill~(Ours)}   &\textbf{62.7}  &51.7 & \textbf{66.5} & 55.0 &59.0 & \textbf{50.5}  & \textbf{54.4} \\
\textbf{\textbf{SPLC}~(Ours)}   &\textbf{62.9} &61.7 &56.8  &\textbf{55.4}  &71.8 & 38.2  & 49.9 \\

\hline

\end{tabular}
\label{tab:app_openimagesingle}
}
\end{center}
\end{table*}

\end{document}